%% file: main.tex
\documentclass[10pt,twocolumn,letterpaper]{article}

\usepackage{cvpr}              

\usepackage[accsupp]{axessibility}

\input{preamble}

\usepackage[pagebackref,breaklinks,colorlinks,citecolor=cvprblue]{hyperref}

\allowdisplaybreaks[1] 

\def\paperID{15680} 
\def\confName{CVPR}
\def\confYear{2024}

\title{Soften to Defend: Towards Adversarial Robustness via \\ Self-Guided Label Refinement}

\author{Daiwei Yu$^{1,*}$ \quad Zhuorong Li$^{1, *, \dagger}$ \quad Lina Wei$^{1}$ \quad Canghong Jin$^{1}$ \quad Yun Zhang$^{1}$ \quad Sixian Chan$^{2}$\\
$^{1}$Hangzhou City University \quad $^{2}$Zhejiang University of Technology\\
{\tt\small $\{$lizr, weiln, jinch, yunzhang$\}$@hzcu.edu.cn, ydw.ccm@gmail.com, sxchan@zjut.edu.cn}}

\begin{document}
\maketitle
\input{sec/0_abstract}    

\input{sec/1_intro}

\input{sec/2_method}

\input{sec/3_exp}

\section*{Acknowledgement}
This work is partially supported by the National Science and Technology Major Project of China (Grant No. 2022ZD0119103), the National Natural Science Foundation of China (Grant No. 61906168), the Zhejiang Provincial Natural Science Foundation of China (Grant No.LQ21F020006, LY23F020023),
and is also supported by the advanced computing resources provided by the Supercomputing Center of Hangzhou City University.

{
    \small
    \bibliographystyle{ieeenat_fullname}
    \bibliography{ref}
}

\input{sec/X_suppl}

\end{document}

%% file: preamble.tex
\usepackage{microtype}              
\usepackage{nicefrac}  
\usepackage{comment}
\usepackage{multirow}
\usepackage{colortbl} 

\usepackage{algorithm}
\usepackage{algorithmic}

\usepackage{xcolor}
\definecolor{cvprblue}{rgb}{0.21,0.49,0.74}
\definecolor{eqpurple}{RGB}{217,217,255}
\definecolor{eqorange}{RGB}{255,217,179}
\definecolor{emph}{RGB}{229, 229, 229}
\definecolor{comparison}{RGB}{29,191,28}

\newtheorem{theorem}{Theorem}
\newtheorem{pro}{\bf Proposition.}

\newtheorem{definition}{\bf Definition.}
\newtheorem{lemma}{\bf Lemma.}

\usepackage{pifont}
\newcommand{\xmark}{\ding{55}}
\newcommand{\cmark}{\ding{51}}

\newcommand{\Tstrut}{\rule{0pt}{2.6ex}}
\newcommand{\Bstrut}{\rule[-0.9ex]{0pt}{0pt}}
\newcommand{\TBstrut}{\Tstrut\Bstrut}

\newcommand\qedd{\par \hfill $\blacksquare$ \par}

\newcommand\blfootnote[1]{%
  \begingroup
  \renewcommand\thefootnote{}\footnote{#1}%
  \addtocounter{footnote}{-1}%
  \endgroup
}

%% file: sec/0_abstract.tex
\begin{abstract}
  \blfootnote{$^\dagger$ Corresponding author.}
  \blfootnote{$^*$ The first two authors contribute equally.}
  Adversarial training (AT) is currently one of the most effective ways to obtain the robustness of deep neural networks against adversarial attacks. 
  However, most AT methods suffer from robust overfitting, i.e., a significant generalization gap in adversarial robustness between the training and testing curves. 
  In this paper, we first identify a connection between robust overfitting and the excessive memorization of noisy labels in AT from a view of gradient norm. 
  As such label noise is mainly caused by a distribution mismatch and improper label assignments, we are motivated to propose a label refinement approach for AT. 
  Specifically, our Self-Guided Label Refinement first self-refines a more accurate and informative label distribution from over-confident hard labels, and then it calibrates the training by dynamically incorporating knowledge from self-distilled models into the current model and thus requiring no external teachers. 
  Empirical results demonstrate that our method can simultaneously boost the standard accuracy and robust performance across multiple benchmark datasets, attack types, and architectures. 
  In addition, we also provide a set of analyses from the perspectives of information theory to dive into our method and suggest the importance of soft labels for robust generalization.
\end{abstract}

%% file: sec/1_intro.tex
\setlength{\parskip}{0pt plus0pt minus0pt}

\section{Introduction}

Recent studies have reported that deep neural networks (DNNs) are vulnerable to adversarial examples, \textit{i.e.}, malicious inputs perturbed by an imperceptible noise to confuse the classifier prediction \cite{explain,intriguing}. This vulnerability raises serious security concerns and motivates a growing body of works on defense techniques \cite{madry,trades,nuat}. 
Adversarial training (AT) is arguably the most promising way to harden classifiers against adversarial examples, which directly augments the training set with adversarial examples \cite{false,madry}. It is formulated as a min-max problem \cite{madry} to find model parameters $w$ that minimize the adversarial risk:
\begin{equation}
	\min _w \mathcal{L}_\mathcal{S}(x^\prime,y;w)
\end{equation}
where $\mathcal{L}_{\mathcal{S}}(x^\prime,y;w)= \nicefrac{1}{n} \sum _{i=1}^{\vert \mathcal{S} \vert} \max \limits_{x^\prime \in \mathcal{B}_r(x)} \ell(f(x_i^\prime;w),y_i)$ and $f(\cdot;w)$ is a model parameterized by $w$, $\ell(\cdot)$ is the loss function such as cross-entropy loss, and $\mathcal{B}_{r}(x)$ denotes the set of the allowed perturbations under the given metric space $M=(X,d)$ and the suitable radius $r>0$, \textit{i.e.}, $\mathcal{B}_r(x)= \{ x+\delta \in \mathcal{X}: d(x,x+\delta)<r \}$.

However, most AT methods suffer from a dominant phenomenon that is referred to as “robust overfitting”.
That is, an adversarially trained model can reach almost 100\% robust accuracy on the training set while the performance on the test set is much inferior, witnessing a significant gap of adversarial robustness~\cite{rice}.
Various regularization techniques including classic $\ell_1$, $\ell_2$ regularization and more advanced regularizations using data augmentation, such as Mixup~\cite{mixup} and Cutout~\cite{cutout}, have been attempted to mitigate robust overfitting, whereas they are reported to perform no better than a simple early stopping~\cite{rice}.
However, early stopping raises another concern as the checkpoint of the best robustness and that of the best standard accuracy often do not coincide~\cite{swa}.
To outperform data augmentations and early stopping, regularizations specifically designed for robust training are thus proposed, to name a few, loss reweighting~\cite{mart,DBLP:conf/iclr/ZhangZ00SK21,DBLP:conf/nips/WangLHLGNZS21,mlcat} and weight smoothing~\cite{swa,awp,DBLP:conf/ijcai/Yu0GSGBL22,mlcat}.
These regularization methods are likely to restrict the change of training loss by suppressing perturbations with respect to either the inputs $x$ or the weights $w$, whereas few explorations attempt to combat robust overfittiing from the perspective of labels $y$. 

\begin{figure*}[t]
	\centering
	\includegraphics[width = \linewidth]{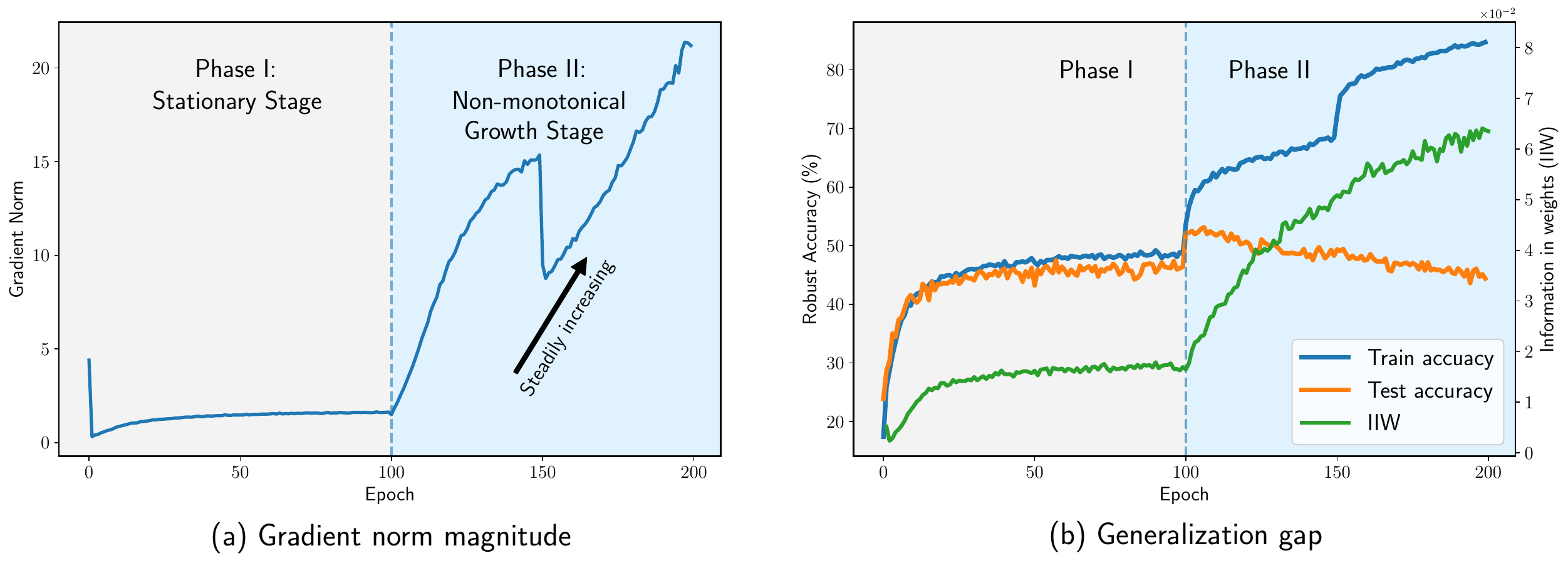}
	\vspace{-2em}
	\caption{In figure (a), we calculate the gradient norm of vanilla adversarially trained PreAct-ResNet 18 on CIFAR-10 for robustness against $\ell_\infty$ perturbations of radius $8/255$. In figure (b), we show the robust accuracy under PGD-20 attack under the same settings with figure (a).
	The gradient norm keeps non-monotonically ramping up when robust overfitting happens.}
	\label{fig:norm}
	\vspace{-1.5em}
\end{figure*}

\begin{figure*}[t]
	\centering	
	\includegraphics[width=\textwidth]{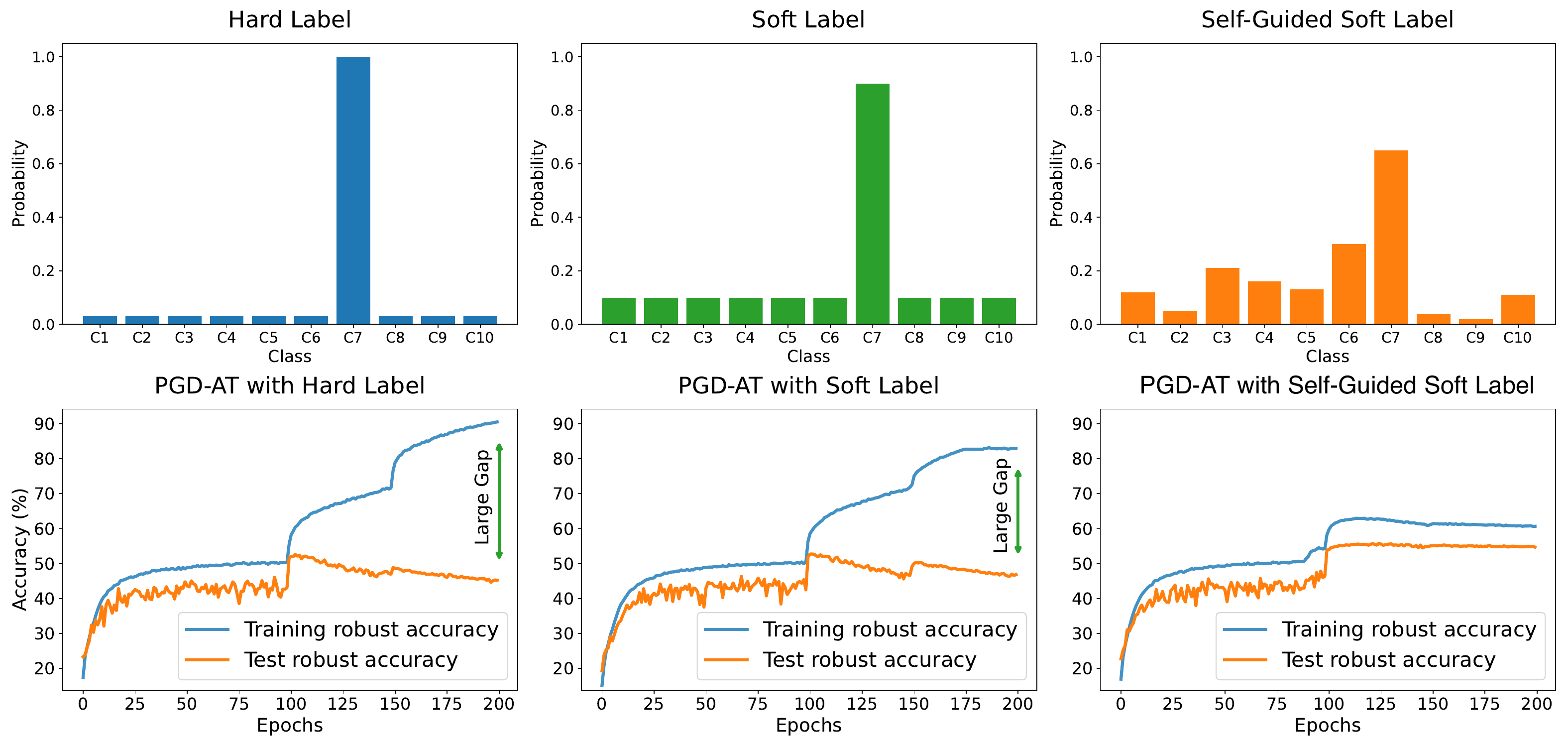}
	\vspace{-2em}
	\caption{Robust accuracy of models employing different label assignment methods in adversarial training.}
	\label{fig:ro}
	\vspace{-1.5em}
\end{figure*}

Previous investigations on labels in AT~\cite{DBLP:conf/icml/YangYYSM20,kd_at} emphasised the existence of label noise in AT to provide an understanding of robust generalization, for instance, from a classic bias-variance perspective.
We take a step still further to investigate the memorization of label noise in AT, which is characterised by a non-monotonical increase of gradient norm, as illustrated in~\cref{fig:norm}, resulting in robust overfitting.
In light of our analyses, we are motivated to design a strategy for label refinement to alleviate excessive memorization and thus the robust overfitting.
To that end, we conduct an empirical experiment using different label assignments and analyse through the lens of learning curve, which is a useful indicator for the occurrence of robust overfitting.
As shown in \cref{fig:ro}, the phenomenon of robust overfitting is presented, for instance, in the lower left of \cref{fig:ro} for PGD-AT \cite{madry} with commonly used hard labels, and at a slightly reduced extent as shown in the lower middle of \cref{fig:ro} with vanilla soft labels.\looseness=-1

In this work, we propose a theoretical-grounded method to alleviate the memoization on over-confident labels during adversarial training and thus to combat robust overfitting.
Our main idea is to resort to an alternative for label assignment.
Particularly, motivated by the effectiveness of soft label in alleviating overﬁtting in standard training \cite{ls}, our work generates more reliable labels automatically by incorporating predictive label distributions into the process of robust learning.
It provides a promising way to inject distilled knowledge and to self-calibrate the adversarial training.
Our method is conceptually simple yet significantly enhances the learning of deep models in adversarial scenarios.
The key contributions are as follows:

\begin{itemize}[leftmargin=12pt]
	\item We first inspect the behaviour of deep models trained by AT when robust overfitting occurs. Specifically, we identify a connection between robust overfitting and the excessive memorization of noisy labels in AT through the lens of gradient norm (see \cref{fig:norm}). As such label noise is mainly due to unaccurate label distribution, we further investigate the effects of different label assingment methods on AT (see \cref{fig:ro}). These observations consistently implies a connection between robust overfitting and noisy hard labels. Exploring such connections could help to shed light on understanding and improving the robust learning of deep models. 
	\item Upon the observation, we are motivated to propose a label refinement approach for AT. Specifically, our Self-Guided Label Refinement (SGLR) first self-refines accurate and informative label distribution from the over-confident hard labels, and then it calibrates the training by dynamically incorporating knowledge from self-distilled models into the current model, requiring no external teacher models nor modifications to the existing architecture.
	\item We verify the effectiveness of SGLR through extensive evaluations. Overall, experimental results show that the proposed method consistently improves the test accuracy over the state-of-the-art on various benchmark datasets against diverse adversaries. Moreover, our approach can achieve robust accuracy up to 56.4\% and close the generalization gap to merely 0.4\%, significantly mitigating the overfitting issue and thus being able to pushing up the adversarial robustness.
\end{itemize}

\section{A closer look at robust overfitting}\label{sec:2}


Robust overfitting has been prevalent across various datasets and models~\cite{rice}. Then, crops of empirical and theoretical studies have emerged to analyze this phenomenon through the lens of \textit{{loss value}}~\cite{mlcat}, \textit{{training data}}~\cite{kd_at} and \textit{{learned features}}~\cite{wang2023balance}. However, a comprehensive underlying mechanism of robust overfitting still remains an enigma. In this section, we revisit robust overfitting from a perspective of the established information theory and identify a “{\textit{memorization effect}}” over the course of adversarial training, which is akin to standard training as oberved in~\cite{DBLP:conf/iclr/XiaL00WGC21,DBLP:conf/iclr/0012ZSL23}. 
First and foremost, we conduct some observations when robust overfitting occurs via analysing the change of the gradient norm of adversarial loss with respect to model weight, \textit{i.e.}, $\Vert \nabla_w L(x^\prime,y,w) \Vert_2$. As shown in \cref{fig:norm} (a), we may note that the gradient norm holds nearly the constant and then keeps ramping up non-monotonically. This increasing behavior of gradient norm encountered with the learning rate (LR) decays. So we divide the training process into two phases according to the LR decays: (1) \textbf{\textit{Stationary stage}}; (2) \textbf{\textit{Non-monotonical growth stage}}.

In stationary stage, as illustrated in \cref{fig:norm} (b), the learning curve (orange line) and test robust accuracy (blue line) ramp up at almost the same pace and maintain a very small generalization gap, \textit{i.e.}, the divergence between the training and test accuracy, implying that the model keeps learning efficient robust features as the training progresses and accrodingly hovers a small constant gradient norm. However, a significantly expanding generalization gap can be witnessed after an inflection point, where the LR decays. We refer the period afterwards as “Non-monotonical growth stage”, as a prominent characteristic of this stage is that the gradient norm keeps growing non-monotonically and does not converge to a constant, even after the training has ended. 

It is worth noting that this trend of gradient norm greatly contradicts with that of the conventional ERM training, where zero gradient norm can be reached at the end of training. Nonetheless, such behaviour of gradient norm is in line with ERM when significant noisy labels surrounding, as observed in~\cite{DBLP:journals/mlst/FengT21,DBLP:conf/iclr/WangM22}. Specifically, such growing trend is interpreted as an indicator that the training has entered a “memorization routine”, where the model firstly fits on training samples with clean labels, then gradually over-fits samples with noisy labels. Later in this stage, the test accuracy on clean data will go down whereas the training accuracy keeps going up. Similarly, we are able to identify a non-monotonical growth in adversarial training, as depicted in \cref{fig:norm} (a). Remarkably, this divergent gradient norm as the adversarial training progressing, exactly accompanies with the enlargement of generalization gap that is depited in \cref{fig:norm} (b), which reveals the overfitting problem. Thus, we come to a conjecture that the memorization effect is also the curprit of the increasing gradient norm in the scenario of robust training and further induces robust overfitting. To that end, we first provide proof for our hypothesis that advesarial training memorizes samples with noisy labels mainly in the non-monotonical divergence phase.

Following~\citep{DBLP:journals/jmlr/AchilleS18}, the expected cross entropy loss could be decomposed into three terms according to PAC-Bayesian framework:
\begin{small}
	\begin{align}
		\begin{split}
			&\mathcal{H}_f(\hat{y} \vert x ,w) 
			\\
			&= \mathbb{E}_\mathcal{S} \mathbb{E}_{w \sim Q(w \vert \mathcal{S})} \sum_{i=1}^{m}
			\left[
			-\log f(\hat{y_i} \vert x_i,w)
			\right] \\
			&= \mathcal{H}(y \vert x) +
			\mathbb{E}_{x, w \sim Q(w \vert \mathcal{S})} \text{KL}[p(y\vert x) \, \Vert \, f(\hat{y}\vert x,w)] - I(w;y \vert x)
		\end{split}
		\label{eq:de_xent}
	\end{align}
\end{small}

In conventional ERM training, minimizing $-I(w;y\vert x)$ relies on the cross entropy loss between the prediction $f(\hat{y} \vert x,w)$ and the true label distribution $p(y\vert x)$.
Noisy label viewed as the outlier of a true label distribution can provide a positive value of $I(w;y\vert x)$.
This term essentially quantifies the extent to which label information overlaps the weights of the model.
In other words, the presence of noisy labels will reduce the finite degree of freedom with respect to weights.
The reason for this reduction is that model attempts to accommodate the noise labels in the training data, inducing a potential overfitting.
Different from the conventional ERM training, adversarial training requires perturbed samples $x^\prime$, which are usually obtained by solving a multi-step maximization problem.
Accordingly, we use the notation $p(y^\prime \vert x^\prime)$ for the true distribution of adversarial samples. Based on the \cref{eq:de_xent}, we have
\begin{small}
	\begin{align}
		\begin{split}
		\mathcal{H}_f(\hat{y} \vert x^\prime ,w) = &\mathbb{E}_\mathcal{S} \mathbb{E}_{w \sim Q(w \vert \mathcal{S})} \sum_{i=1}^{m}
		\left[
		-\log f(\hat{y_i} \vert x_i^\prime,w)
		\right] \\
		= &\mathcal{H}(y^\prime \vert x^\prime) - I(w;y^\prime \vert x^\prime) \\
		+ & \mathbb{E}_{x^\prime, w \sim Q(w \vert \mathcal{S})} \text{KL}[p(y^\prime \vert x^\prime) \, \Vert \, f(\hat{y}\vert x^\prime,w)]
		\end{split}
	\end{align}
\end{small}

As aforementioned, the adversarial perturbation cause a mismatch between the label distributions of the perturbed data and their origins, as $p(y^\prime \vert x^\prime) \neq p(y \vert x)$. However, during adversarial training,  $p(y^\prime \vert x^\prime)$ is often simply replaced by $p(y\vert x^\prime)$, that is, the perturbed labels are directly inherited from their origins. This would inevitably exert influence on the label memorization and results in performance degradation.

We are interested in the memoization of noisy labels in adversarail training, which can be characterised by $I(w;y^\prime\vert x^\prime)$ and refered to as “Information In Weights” (IIW). The training dynamics can also be captured through the lens of IIW. However, computing the value of $I(w;y^\prime\vert x^\prime)$ is as difficult as fitting the model itself. Basing on the chain rule of mutual information, we could approach to our desired result via the upper bound, \textit{i.e.}, $I(\mathcal{D};w) = I((x,y);w) = I(x;w) + I(y;w \vert x)$. By using the positivity property of mutual information, we have $I(y;w\vert x) \leq I(D ; w)$. Then we could take an approximation under some Gaussian assumptions, as suggested by \citet{DBLP:conf/iclr/0008HKSC022}, to estimate $I(w;y^\prime\vert x^\prime)$ and further depict the learning behaviour of model, which is illustrated by the green line in \cref{fig:norm} (b). The increasing IIW supports our hypothesis that adversarially trained model mainly overfits the noisy labels in the non-monotonical growth stage. Upon the theoretical analysis, it is natural that we proposed to prevent models from excessively memorising noisy labels by means of IIW reductions. In this work, we propose to reduce IIW through soft labels, which is specifically effective in lowering mutual information and thus can be expected to weaken the memorization under the distribution mismatch. 
\begin{theorem}[Soft label could reduce the IIW]\label{thm:iiw}
	Let $u$ be the uniform random variable with p.d.f $p(u)$. By using the composition in \cref{eq:de_xent}, there exists an interpolation ration $\lambda$ between the clean label distribution and uniform distribution, such that
	\begin{align}
		I(y^\ast; w\vert x^\prime) \lesssim I(y;w\vert x^\prime)
	\end{align}
	where $p(y^\ast \vert x^\prime , w) = \lambda \cdot p(y \vert x^\prime, w) + (1-\lambda) \cdot p(u)$ and the symbol $\lesssim$ means that the corresponding inequality up to an $c$-independent constant.
\end{theorem}
The detailed proof could be found in \cref{appendix:proofs}. 
\cref{thm:iiw} proves that there exists some kind of soft label that could reduces the information in weights. So the memorization effect caused by the label distribution mismatch could be effectively mitigated. Thus, to better suppress the memorization effect, we should provide more underlying information about the true label distribution than uniform distribution to facilitate a better interpolation of the soft label. In the following, we introduce our solution to estimate more accurate and informative soft labels for adversarial training.

%% file: sec/2_method.tex
\section{Self-guided label refinement for adversarial training}

\subsection{Methodology}
As discussed above, hard labels in adversarial training are uninformative but over-confident, and thus heavily impairs the generalization (see \cref{fig:ro}). To address this issue, we propose an alternative to one-hot labels for adversarial training. Specifically, our Self-Guided Label Refinement first utilizes the learned probability distributions to refine the over-confident hard labels, and then it guides the training process using the historical training model to obtain a calibrated prediction.

We begin by softening the over-confident hard labels. It is well acknowledged that label smoothing (LS) helps to calibrate the degree of confidence of a model and it is effective in improving the robustness in noise regimes \cite{ls_noise}. The vanilla LS for a $K$-class classification problem can be formulated:
\begin{equation}
	\mathbf{y} = \frac{r}{K} \cdot \mathbf{1} + (1-r) \cdot \mathbf{y}_{hard}
\end{equation}
where, $\mathbf{y}_{hard}$ denotes labels encoded by a one-hot vector, the notation of $\mathbf{1}$ denotes all one vector, and $r \in [0,1]$ controls the smooth level. It is shown that LS serves as a regularizer as well as {a confidence penalty \cite{cp} }and therefore improves the generalization of the model in standard training. Unfortunately, when it comes to adversarail training, a direct combination with LS cannot guarantee reliable robustness, especially in the cases of strong perturbations\cite{eva,bag}. Specifically, as the uniform distribution is unlikely to match the underlying distribution, LS tends to introduce a bias that might hurt the robust generalization. Moreover, soft labels with identical probability over the false categories cannot reveal the semantic similarity.

To alleviate this artificial effect, we introduce our Self-Guided Label Refinement (SGLR) which utilizes the knowledge inferred by a trustworthy model itself to retains informative labels. The proposed SGLR can be formulated as:
\begin{equation}
	\mathbf{y} = r \cdot f(x^\prime;w) + (1-r) \cdot \mathbf{y}_{hard}
    \label{eq:2}
\end{equation}
where $f(x^\prime;w)$ is the logit output of the model parameterized by $w$ on training data $x^\prime$. To be noticed, the model $f$ we referred to here is not pre-trained but rather on the training. {\cref{eq:2} also} in fact serves as a regularizer by integrating the knowledge extracted from the model with stastic information that encoded by one-hot labels, thus, we suggest that this particular form of soft label has the potential to convey a better quantity of informative content. 

Furthermore, according to \citet{benign}, there exists a good interpolation of the noisy training dataset that could lead to a satisfactory generalization.
Consequently, we aim to establish a valid interpolation between robust and non-robust features with the expectation that it strikes a balance between accuracy and robustness, as represented by the equation:
\begin{equation}    
	\widetilde{f}(x, x^\prime;w) = \lambda \cdot f(x;w) + (1-\lambda)\cdot    f(x^\prime;w)
	\label{eq:7}
\end{equation}

\begin{figure}[t]
	\centering
	\includegraphics[width = \linewidth]{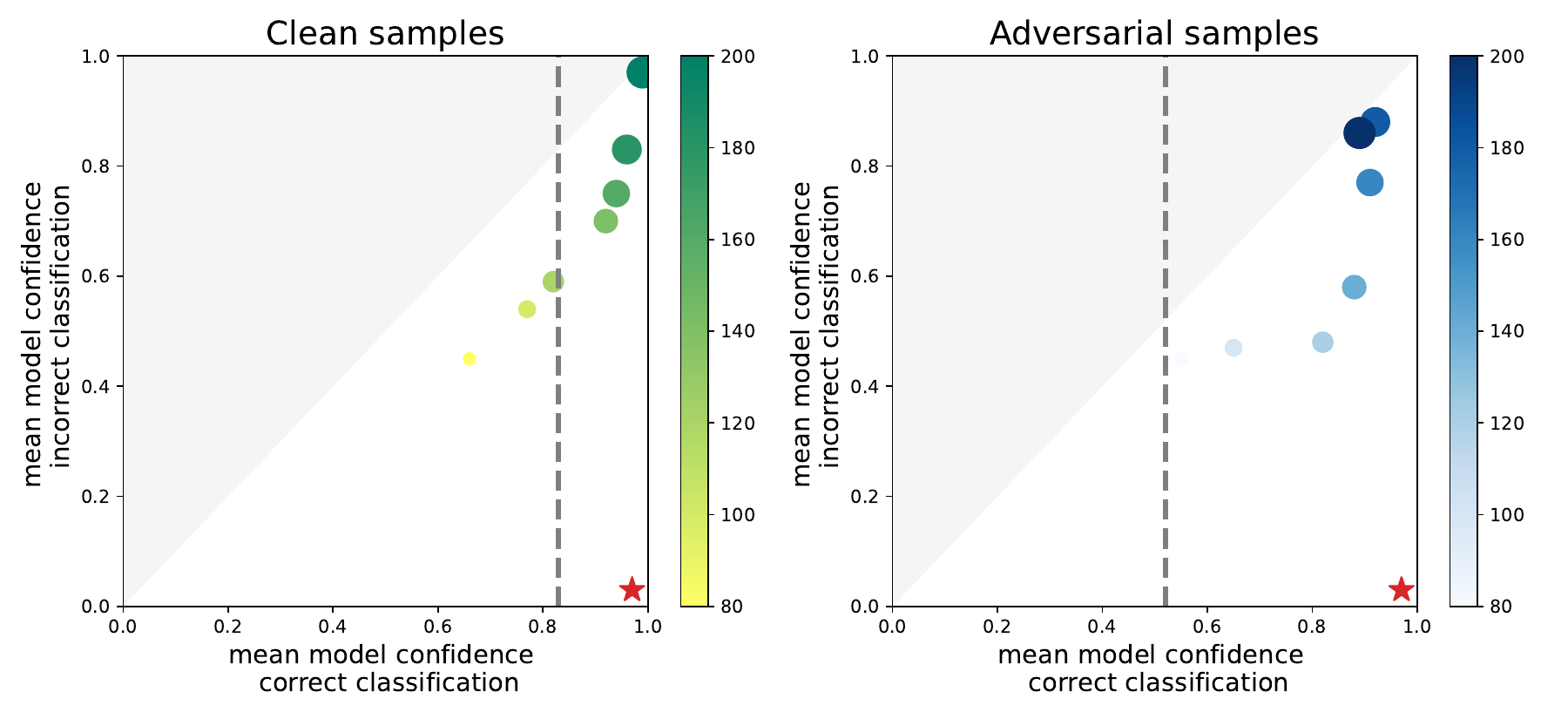}
	\vspace{-2em}
	\caption{The mean confidence of model in the correct and incorrect predictions over clean and adversarial test sets.}
	\label{fig:confidence}
	\vspace{-1.5em}
\end{figure}

\cref{fig:confidence} illustrates the major trajectory of PGD-AT training on CIFAR-10 dataset in terms of the confidence in both the correct and incorrect predictions, where the model dynamics are shown by a series of colored dots with darker colours and larger areas. Gray dashed lines mark the best standard / robust accuracy of the trained model, and the red star in the bottom right corner denotes an ideal model that with a high confidence in correct classification whereas a low one in the incorrect prediction. As training processes, model tends to assign increasing confidence to its correct predictions and also to those wrong predictions, as implied by the moving of the colored dots away from the dash line. In other words, the latter prediction of the model is not better calibrated and the model in the middle training stage could be of great help to reduce the expected calibration error. Therefore, we are motivated to utilize exponential moving average (EMA) that taking the moving average of history model prediction to obtain calibrated soft label. Intuitively, EMA considers recent proposals from the current state while retaining some influence from previous information. This straightforward method is easy to deploy and incurs negligible training overhead. Then the dynamic updating of soft label could be defined as follows:
\begin{align}    
	\begin{split}    
		&\mathbf{y} = r \cdot \widetilde{p_t} + (1-r) \cdot \mathbf{y}_{hard} \\    \widetilde{p_t} = &\alpha \cdot \widetilde{p}_{t-1} + (1-\alpha) \cdot \widetilde{f}(x,x^\prime;w_t)    
	\end{split}
\end{align}

\begin{table*}[t]
	\centering
	\caption{Comparison between our method and other KD methods.}
	\vspace{-0.6em}
	\begin{tabular}{c c c c c}
		\toprule[1.5pt]
			& ARD~\cite{ard} & RSLAD~\cite{rslad}  & KD-SWA~\cite{swa}  & Ous \\
		\midrule
		Trained teacher models                     & 1     & 1   & 2   & 0   \\
		Standardly teacher independent                           &   \cmark    &    \cmark  &  \xmark   &  \cmark   \\
		Adversarially teacher independent  & \xmark  & \xmark & \xmark & \cmark  \\
		Forward times in one training iteration   & 3     & 3  & 4     & 2  \\
		\bottomrule[1.5pt]
	\end{tabular}
	\label{tab:ckd}
	\vspace{-0.5em}
\end{table*}

\subsection{Connection to symmetric cross entropy}
Cross Entropy (CE) is the prevalent loss function for deep learning with remarkable success. However, as the noisy label is tipping the training process into overfitting, CE loss has taken a nasty blow and is prone to fitting noise. Inspired by the symmetric Kullback-Leibler (KL) Divergence, \citet{sce} proposed a more flexible loss, \textit{i.e.}, symmetric cross entropy (SCE), to strike a balance between sufficient learning and robustness to noisy labels. Moreover, it reveals that predictive distribution exploited by the model is superior to one-hot label distribution for the most part. Both SCE loss and our method utilize the informative knowledge inferred by the model and \cref{pro:1} provide such a close link between them.
\begin{pro}\label{pro:1}
	Let $\ell_{sce}$ be the SCE loss function and $\gamma$ represents the sum of CE and reverse CE loss. When $\gamma \to 1$, then our methods can also be written as:
	\begin{align*}
		\ell_{sglr} = \ell_{sce} - \alpha \cdot \ell_{rkl}
	\end{align*}
	where $\ell_{rkl}$ denotes the reverse KL divergence between labels and model predictions, \textit{i.e.}, $ D_{\text{KL}}(p \, || \, q)$.
\end{pro}
The proof of \cref{pro:1} can be found in \cref{appendix:proofs}. Note that \cref{pro:1} indicates our method can be decomposed into two terms. The first term represents the SCE loss function when the weighted sum tends to 1. The second term indicates that our method rewards the distribution differences between predictions and labels. Though our study aims at mitigating robust overfitting, which is different from the scenario of SCE loss, there is no conflict between our method and SCE.

\subsection{Comparison to knowledge distillation}\label{comparison:kd}
Knowledge distillation (KD) \cite{kd} has been proven to be an effective way to compress models, which is able to remarkably beef up the performance of lightweight models. There is a huge scope for applying KD to adversarial training to improve the efficiency of AT. A crop of works \cite{ard,mart,rslad,swa} have advanced exploring KD implicitly or explicitly in conjunction with AT. The efficacy of KD is attributed to the teacher model's informative knowledge, which guides the student to acquire a more similar knowledge representation and effectively capture inter-class relationships. By enforcing consistency in probability distributions, KD facilitates better learning outcomes for the student.
\begin{pro}\label{pro:2}
	Some KD methods, which minimize the distance of the feature map between the teacher and student model, belong to the family of our method. Let $p_t$ be the prediction of the teacher model and then the KD could also be written as $\ell_{\text{KD}} = \mathbb{E}_{\widetilde{q}} \, \left[  -\log p  \right] = H(\widetilde{q},\,p)$, where $\widetilde{q} = (1-\alpha) \cdot q + \alpha \cdot p_t$.
\end{pro}

\cref{pro:2} shows that some KD revised the hard label via the knowledge $p_t$ learned by the teacher. The proof of \cref{pro:2} is presented in \cref{appendix:proofs}. Further, we view our method as special supervision, and smoothing the hard label could better reflect the similarities among classes. There are many differences between our method and other KDs in AT. The main differences are summarized in \cref{tab:ckd}. Our method does not incur any extra computational cost as we do not involve teacher models, which is a boon throughout the whole training progress.

\subsection{Tolerant to noisy label}
In the following part, we'd like to delve into whether the proposed method is tolerant of noisy labels. We define the symmetric noise label as that true label $y$ has probabilities $\eta_{x,\widetilde{y}} = p(\widetilde{y}\,|\,y,x)$ to flip into the wrong labels uniformly. The corresponding \textit{noisy empirical risk} is:
\begin{align*}
	R_{\mathcal{S}}^\eta (f) = \mathbb{E}_{\mathcal{S}} (1-\eta_x) \cdot \ell(f(x),y) + \sum_{i \neq y} \eta_{x, i} \ell (f(x),i)
\end{align*}
where, $\eta_x$ is the noise rate. We call a loss function noise-tolerant if and only if the global minimum$f_\eta^\ast$  has the same probability of misclassification as that of $f^\ast$ on the noise-free data.
\begin{lemma}
	Given a symmetric loss function $\ell$ that it satisfies $\sum_{i=1}^k L(f(x),i)=C$ and $C$ is some constant. Then $\ell$ is noise tolerant under symmetric label noise if noise rate $\eta$ meets $\eta < 1 - \frac{1}{K}$.
\end{lemma}

The loss condition theoretically guarantees the noise tolerance by risk minimization on a symmetric loss function following \cite{noise} and it shows that the global optimal classifier $f^\ast$ on noise-free data remains the optimal even with the noisy label. Further, we can also derive the noise tolerance theoretically about our method from \cref{thm:1}.
\begin{theorem}\label{thm:1}
	In a $K$-class classification problem, our method $\widetilde{\ell}$ is noise-tolerant under symmetric or uniform label noise if noise rate $\eta < 1- \frac{1}{K}$. And if $R(f^\ast)=0$, $\widetilde{l}$ is also noise-tolerant under asymmetric or class-dependent label noise when noise rate $\eta_{y,k} < 1- \eta_y$ with $\sum_{i \neq y} \eta_{y, i}=\eta_y$, then
	\begin{align*}
		R_\mathcal{S}^\eta(f^\ast)-  R_\mathcal{S}^\eta(f) \simeq(1 - \frac{\eta K}{K-1})(R_\mathcal{S}(f^\ast) - R_\mathcal{S}(f)) \leq 0
	\end{align*}
\end{theorem}
From \cref{thm:1}, we can derive that our method is nearly noise-tolerant under symmetric noise and further prove the robustness of the proposed method to asymmetric noise. More details can be found in \cref{appendix:results}. We show a significant improvement under noise regimes empirically in experiments.\looseness=-1

%% file: sec/3_exp.tex
\section{Experiments}

\paragraph{Training and evaluation setups.} We conduct extensive experiments on the benchmark datasets, CIFAR-10/100. We set the perturbation budget to $\epsilon=8/255$ under the $\ell_\infty$ norm-bounded constraint. We use ResNet-18 \cite{resnet} as our default network architecture unless otherwise specified. For all experiments, the model is trained using SGD with a momentum of 0.9, weight decay of $5 \times 10^{-4}$, and an initial learning rate of 0.1 for a total of 200 epochs. The learning rate is decayed by a factor of $0.1$ at the $100$-th and $150$-th epochs, following \cite{rice}. The evaluation of the proposed approach encompasses PGD-20 \cite{rice} and AutoAttack \cite{aa}, which is recognized as the most reliable robustness evaluation up to date. AutoAttack is an ensemble of diverse attacks, including APGD-CE, APGD-DLR, FAB \cite{fab} and Square attack \cite{square}. We quantify the robust generalization by computing the difference between the best and final checkpoints over the course of training.

\begin{table*}[t]
	\centering
	\caption{Test accuracy (\%) of the proposed method and other methods on CIFAR-10 under the $\ell_\infty$ norm with $\epsilon= 8/255$ based on the ResNet-18 architecture.}
	\vspace{-0.6em}
	\begin{tabular}{l c c c c c c c c c}
		\toprule[1.5pt]
		\multicolumn{1}{c}{\multirow{2}{*}{Method}} & \multicolumn{3}{c}{Natural Accuracy} & \multicolumn{3}{c}{PGD-20} & \multicolumn{3}{c}{AutoAttack} \\
		\multicolumn{1}{c}{}                        & Best       & Final       & Diff $\downarrow $     & Best    & Final   & Diff $\downarrow $  & Best     & Final     & Diff $\downarrow $   \\
		\midrule[1pt]
		PGD-AT                                          &     80.7       &     82.4        &     -1.6      &    50.7     &   41.4      &   9.3     &   47.7       &    40.2       &   7.5     \\
		PGD-AT+LS                                   &   82.2         &    84.3         &     -2.1      &    53.7     &  48.9       &   4.8     &   48.4       &    44.6       &   3.9      \\
		PGD-AT+TE                                   &    82.4        &    82.8         &    -0.4       &   55.8      &   54.8      &    1.0    &    50.6      &    49.6       &    1.0 \\ PGD-AT+SGLR                                    &   82.9       &       83.0     &    \textbf{-0.1}       &   \textbf{56.4}      &    \textbf{55.9}      &   \textbf{0.5}     &    \textbf{51.2}      &     \textbf{50.2}      &    1.0     
		\\ \hline \TBstrut
		AWP                                      &   82.1         &   81.1          &   1.0        &   55.4   &   54.8    &    0.6    &    50.6      &      49.9     &   0.7      \\
		KD-AT                                      &   82.9         &   \textbf{85.5}          &   -2.6        &   54.6   &   53.2    &    1.4    &    49.1      &      48.8     &   0.3      \\
		KD-SWA                                      &   \textbf{84.7}         &   85.4          &   -0.8        &   54.9   &   53.8    &    1.1    &    49.3      &    49.4       &   \textbf{-0.1}      \\
		PGD-AT + SGLR                                      &   82.9       &       83.0     &    \textbf{-0.1}       &   \textbf{56.4}      &    \textbf{55.9}      &   \textbf{0.5}     &    \textbf{51.2}      &     \textbf{50.2}      &    1.0     \\
		\bottomrule[1.5pt]
	\end{tabular}
	\label{tab:3}
	\vspace{-0.5em}
\end{table*}  

\begin{table*}[t]
	\centering
	\caption{Clean accuracy and robust accuracy (\%) of ResNet 18 trained on different benchmark datasets. All threat models are under $\ell_\infty$ norm with $\epsilon=8/255$. The bold indicates the improved performance achieved by the proposed method.}
	\vspace{-0.6em}
	\begin{tabular}{c l c c c l l l l l l}
		\toprule[1.5pt]
		\multirow{2}{*}{Dataset}   & \multicolumn{1}{c}{\multirow{2}{*}{Method}} & \multicolumn{3}{c}{Natrural Accuracy} & \multicolumn{3}{c}{PGD-20} & \multicolumn{3}{c}{AutoAttack} \\
		& \multicolumn{1}{c}{}                        & Best       & Final       & Diff $\downarrow $      & Best    & Final   & Diff $\downarrow $  & Best     & Final     & Diff $\downarrow $    \\
		\midrule[1pt]
		\multirow{4}{*}{CIFAR-10}  & AT                                          &  80.7       &     82.4        &     -1.6      &    50.7     &   41.4      &   9.3     &   47.7       &    40.2       &   7.5        \\
		& \cellcolor{emph}{\textbf{+SGLR}}                                 &   \cellcolor{emph}{\textbf{82.9}}         &    \cellcolor{emph}{\textbf{83.0}}           &   \cellcolor{emph}{\textbf{0.1}}           &   \cellcolor{emph}{\textbf{56.4}}        &    \cellcolor{emph}{\textbf{55.9}}       &   \cellcolor{emph}{\textbf{0.5}}       &    \cellcolor{emph}{\textbf{51.2}}        &       \cellcolor{emph}{\textbf{50.2}}      &   \cellcolor{emph}{\textbf{1.0}}        \\ \cline{2-11} \TBstrut
		& TRADES                                      &  81.2          &     82.5        &     -1.3       &    53.3     &  50.3       &   3.0     &    49.0      &    46.8       &   2.2      \\
		& \cellcolor{emph}{\textbf{+SGLR}}                                 &   \cellcolor{emph}{\textbf{82.2}}         &    \cellcolor{emph}{\textbf{83.3}}           &   \cellcolor{emph}{\textbf{-0.9}}           &   \cellcolor{emph}{\textbf{55.8}}        &    \cellcolor{emph}{\textbf{55.4}}       &   \cellcolor{emph}{\textbf{0.4}}       &    \cellcolor{emph}{\textbf{50.7}}        &       \cellcolor{emph}{\textbf{50.1}}      &   \cellcolor{emph}{\textbf{0.6}}        \\
		\hline \TBstrut
		\multirow{4}{*}{CIFAR-100} & AT                                          &     53.9        &    53.6         &   0.3        &    27.3     &   19.8      &   7.5     &   22.7       &       18.1    &     4.6      \\
		& \cellcolor{emph}{\textbf{+SGLR}}                                 &   \cellcolor{emph}{\textbf{56.9}}         &    \cellcolor{emph}{\textbf{56.6}}           &   \cellcolor{emph}{\textbf{0.3}}           &   \cellcolor{emph}{\textbf{34.5}}        &    \cellcolor{emph}{\textbf{34.3}}       &   \cellcolor{emph}{\textbf{0.2}}       &    \cellcolor{emph}{\textbf{27.5}}        &       \cellcolor{emph}{\textbf{26.7}}      &   \cellcolor{emph}{\textbf{0.8}}        \\
		\cline{2-11} \TBstrut
		& TRADES                                      &    \textbf{57.9}       &     56.3        &   1.7         &   29.9      &  27.7       &   2.2     &   24.6       &    23.4       &   1.2      \\
		& \cellcolor{emph}{\textbf{+SGLR}}                                 &   \cellcolor{emph}{57.1}         &    \cellcolor{emph}{\textbf{57.4}}           &   \cellcolor{emph}{\textbf{-0.3}}           &   \cellcolor{emph}{\textbf{33.9}}        &    \cellcolor{emph}{\textbf{33.2}}       &   \cellcolor{emph}{\textbf{0.7}}       &    \cellcolor{emph}{\textbf{27.1}}        &       \cellcolor{emph}{\textbf{26.4}}      &   \cellcolor{emph}{\textbf{0.7}}        \\
		\bottomrule[1.5pt]       
	\end{tabular}
    \label{tab:2}
	\vspace{-0.5em}
\end{table*}

\begin{figure}[b]
	\centering
	\includegraphics[width=\linewidth]{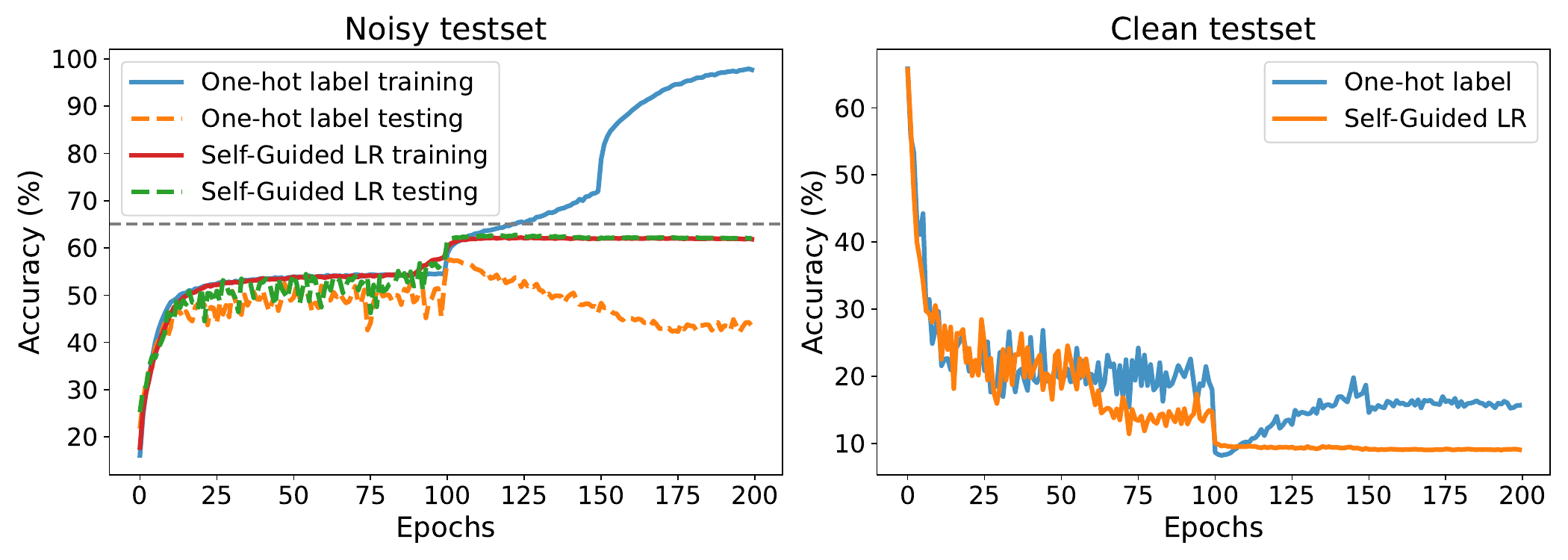}
	\vspace{-2em}
	\caption{Test accuracy (\%) on CIFAR-10 dataset (with 40\% label noise). We split the training set into 1) \textit{untouched portion}, where the labels of elements are left untouched; 2) \textit{corrupted portion}, where the labels of elements are assigned uniformly at random.}
	\label{fig:4}
	\vspace{-1em}
\end{figure}

\vspace{-0.5em}
\paragraph{Improved robust performance across datasets.}
Experimental results of PGD-AT \cite{madry}, TRADES \cite{trades}, and the combinations of them with our proposal (denoted as PGD-AT+SGLR and TRADES+SGLR) on CIFAR-10/100 datasets are shown in \cref{tab:2}, from which we make the following observations on our advantages: 1) \textbf{Closer generalization gap}. It is remarkable that the differences between the best and the final test accuracies of the combinations are reduced to around 0.5\%, while the corresponding baselines (PGD-AT and TRADES) induce much larger gaps, which are up to 9.3\% and 3.0\% on CIFAR-10 dataset when the model is threaten by PGD-20.  It indicates that our method effectively alleviates robust overfitting. 2) \textbf{Higher robust accuracy}. As the combinations induce smaller gaps, they can approach higher robust accuracy against adversaries compared to baseline methods, as can be observed in \cref{tab:2}. Besides, it is notable that our method is especially effective against AutoAttack \cite{aa}, regarding the fundamental difficulties of simultaneously achieving the robustness against multiple adversaries. 3) \textbf{Consistent improvement}. Combining our label smoothing method could consistently and significantly improve the performance on all considered adversaries and also on the clean test set, and across different datasets, namely, the CIFAR-10 and  CIFAR-100 datasets.

\vspace{-0.5em}
\paragraph{Improved generalization under noise regimes.}
Moreover, we verify the self-guided label smoothing by comparing the behavior of the proposed SGLR w
ith that of the widely used one-hot labels under different noise settings, with $40\%$ noisy labels and fully $60\%$ true labels, respectively. Specifically, we focus on evaluating the effectiveness of our approach in mitigating the impact of \textit{symmetric} noisy labels, where the labels are resampled from a uniform distribution over all labels with a probability $\eta$. 

We begin by making the observations in the left of \cref{fig:4}: 1) The generalization error (\textit{i.e.}, the difference between the training and test accuracy) of the proposed SGLR (almost negligible) is obviously smaller than that of the commonly used hard labels ($>50\%$). 2) The peak of the training accuracy using hard labels (blue curve) can approximate 100\% even on the training set with $40\%$ noisy labels (horizontal gray dashed denotes the portion of correct labels), which suggests that using hard labels could fit correct labels in the early stage and eventually memorizes noisy labels. On the contrary, our training curve is bounded by the untouched portion, implying that our method is able to calibrate the training process and maintain a proper fitting of training data. Then we turn to the other setting as shown in the right of \cref{fig:4}, our test accuracy grows steadily while the baseline method fails. This observation again suggests that the double-descent phenomenon might be due to the overfitting of noise and it is hopeful to be avoided by our method. 

\begin{figure}[t]
	\centering
	\includegraphics[width=\linewidth]{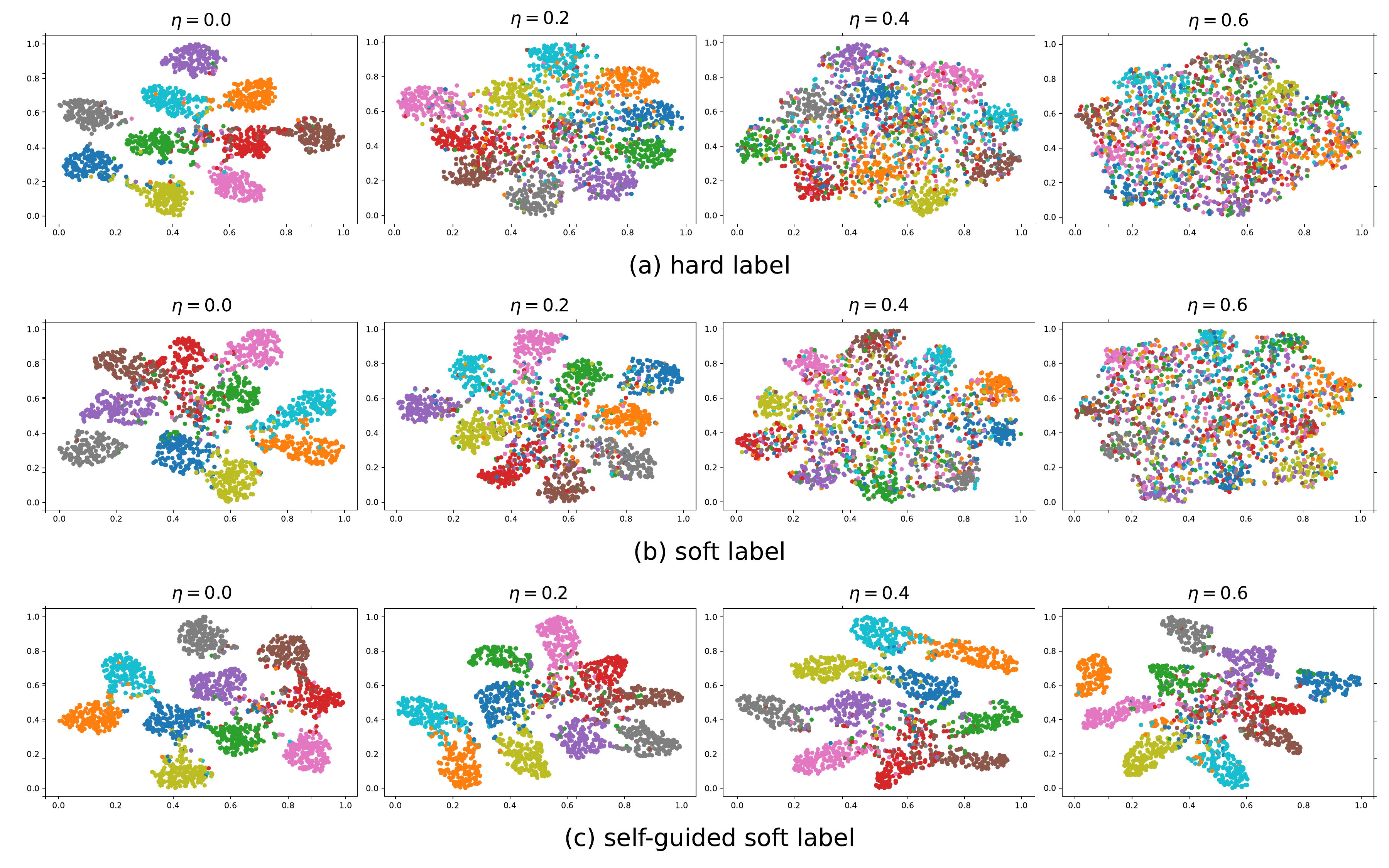}
	\vspace{-2em}
	\caption{Visualization of representations learned by standard training with hard/soft labels and the proposed SGLR with self-guided distribution on CIFAR-10 dataset under various levels of symmetric noisy labels ($\eta \in [ 0.0, 0.2, 0.4, 0.6 ]$).}
	\label{fig:tsne}
	\vspace{-1.5em}
\end{figure}

To give a more intuitive explanation, we further visualize the penultimate layer representations of ResNet 18 trained with (a) a hard label; (b) a soft label, and (c) the proposed soften label. As \cref{fig:tsne} shown, under all the settings of noise rate, the proposed SGLR with self-guided distribution can consistently provide better representations, displayed with more separable clusters and tighter intra-class distance, which are very close to those of the clean setting (the leftmost column). Notably, even with a significantly increased noise rate from 0 (no noise) to 0.6 (severe noise), as illustrated from left to right in \cref{fig:tsne}, our method learns representations with almost negligible variance. This also echoes our advantages of robust learning under various settings of data corruption.

In a nutshell, we have grounds to empirically believe that our method exhibits benign tolerance to noisy labels and improves the generalization ability of the model by avoiding fitting noisy labels. 

\vspace{-0.5em}
\paragraph{Comparison against other methods.}
Considering that our method is analogous to employing the technique of label smoothing into AT, in \cref{tab:3}, we report robustness evaluations of AT with label smoothing (PGD-AT + LS) and temporal ensembling (PGD-AT + TE) on CIFAR-10 test set. 
Since excessive LS could degrade the robustness reported in \cite{bag}, we implement LS with smoothing level $r=0.2$. 
We empirically found that SGLR not only can effectively alleviate robust overfitting while LS fails but also boosts the robustness even under strong AutoAttack. 
Further, we also report other methods (\textit{e.g.}, AWP \cite{awp}, KD-AT \cite{kd_at}) in \cref{tab:3}, which can mitigate robust overfitting through the lens of weight smoothness and the knowledge transfer of teacher model. 
Though robust overfitting is indeed impeded by applying these methods, our method could better narrow the generalization gap between best and final models and achieve remarkable robustness than others under AutoAttack. \looseness=-1

\vspace{-0.5em}
\paragraph{On the impact of smoothing effect $r$.}
The effectiveness of SGLR relies on the smoothing level of soft labels. Besides, as we discussed in \cref{comparison:kd}, SGLR could be viewed as a kind of knowledge distillation without extra teacher models involved. Large temperature $T$ during distillation improves the smoothness of output distribution but could impair the test performance.
As the temperature is vital, we could specify $f(x;w)_i$ in \cref{eq:7} as $\exp(z_i / T) / \sum_{j} \exp (z_j/T)$. We vary the smoothing level $r \in \{0.0,0.2,0.4,0.6,0.8\}$ in \cref{tab:ablation} and note that an increase in $r$ initially boosts both standard accuracy and robust accuracy, followed by a decline,
indicating that excessive smoothing introduces noise and hampers predictive capability. Additionally, increasing the temperature while fixing $r$ only gains slightly improvement in standard accuracy but oberves degradation in robust accuracy.

\begin{table}[h]
	\centering
	\caption{Ablation study on smoothing level $r$ and temperature $T$.}
	\vspace{-0.6em}
	\begin{tabular}{c c c c c c c}
	\toprule[1.5pt]
	{$T / r$} &                 & 0.0   & 0.2   & 0.4   & 0.6   & 0.8 \TBstrut   \\
	\midrule
	\multirow{2}{*}{1}  & SA & 82.4  & 83.5 & 82.8 & 78.5 & 69.1 \\
	& RA  & 43.7 & 54.9 & 54.2 & 52.7 & 48.3 \\
	\midrule
	\multirow{2}{*}{1.5}  & SA & 83.2  & 82.9 & 83.1 & 80.6 & 74.5 \\
	& RA  & 45.7 & 55.9 & 53.2 & 50.4 & 49.8 \\
	\midrule
	\multirow{2}{*}{2}  & SA & 82.9  & 84.4 & 83.9 & 78.5 & 75.6 \\
	& RA  & 45.8 & 53.9 & 52.0 & 50.9 & 47.8 \\
	\bottomrule[1.5pt]
	\end{tabular}
	\label{tab:ablation}
  \end{table}

\section{Discussion and conclusion}
In this study, we show that label noise induced by distribution mismatch and improper label assignments would degrade the test accuracies as well as make the robust overfitting aggravated. From the view of this observation, a label assignment approch for AT, Self-Guided Label Refinement (SGLR), is proposed to weaken the memorization in AT on noisy labels and thus to mitigate the robust overfitting. Extensive experimental results demonstrate the effectiveness of the proposed SGLR. Though we circumvent over confident prediction, the model is still not well-calibrated. As we measure the calibration of model during the training over the entire dataset instead of in a sample-wise way, it may give a false sense of reweighting confidence. In the future, we will attempt to address this limitation.

%% file: sec/X_suppl.tex
\clearpage
\appendix

\setcounter{page}{1}
\maketitlesupplementary

\setcounter{theorem}{0}
\setcounter{pro}{0}
\setcounter{lemma}{0}
\counterwithin{figure}{section}
\counterwithin{table}{section}
\counterwithin{equation}{section}

\section{Proofs}\label{appendix:proofs}

\subsection{Proof of \cref{thm:iiw}}
\begin{theorem}[Soft label could reduce the IIW]
	Let $u$ be the uniform random variable with p.d.f $p(u)$. By using the composition in \cref{eq:de_xent}, there exists an interpolation ration $\lambda$ between the clean label distribution and uniform distribution, such that
	\begin{align}
		I(y^\ast; w\vert x^\prime) \lesssim I(y;w\vert x^\prime)
	\end{align}
	where $p(y^\ast \vert x^\prime , w) = \lambda \cdot p(y \vert x^\prime, w) + (1-\lambda) \cdot p(u)$ and the symbol $\lesssim$ means that the corresponding inequality up to an $c$-independent constant.
\end{theorem}
\textbf{\textit{Proof.}}\,
For this proof, we will use an inequality called the \textit{log-sum inequality}.
\begin{lemma}[Log-sum inequality]
	Let $a_1, \cdots, a_n$ and $b_1, \cdots, b_n$ be nonnegative numbers. Denote the sum of all $a_i$s by $a$ and the sum of all $b_i$s by $b$. The log sum inequality states that
	\begin{align*}
		\sum_{i=1}^n a_i \log \frac{a_i}{b_i} \geq a \log \frac{a}{b}.
	\end{align*}
	with equality if and only if $\dfrac{a_i}{b_i}$ are equal for all $i$.
\end{lemma}
We rewrite the interpolation for simplicity
\begin{align}
	\begin{split}
	p(y^\ast \vert x^\prime , w) &= \lambda \cdot p(y \vert x^\prime , w) + (1-\lambda) \cdot p(u) \\
	p^\ast &= \lambda \cdot p_1 + (1-\lambda) \cdot p_2 \\
	p &= \lambda \cdot p_1 + (1-\lambda) \cdot p_1
\end{split}
\end{align}
Then we could derive the decomposition of cross entropy on different label distribution, i.e., $p^\ast$ and $p$.
\begin{align}
	\mathcal{H}(p^\ast, f) &= \mathcal{H}(p^\ast) - I(w;p^\ast) + \mathbb{E}_{w \sim Q(w \vert \mathcal{S})} \text{KL}[ p^\ast \Vert f ] \notag \\
	&= \lambda \cdot \mathcal{H}(p_1) + (1-\lambda) \cdot \mathcal{H}(p_2) - I(w;p^\ast) + \mathbb{E}^\ast \notag  \\
	&= \lambda \cdot \mathcal{H}(p) + (1-\lambda) \cdot \mathcal{H}(p_2) - I(w;p^\ast) + \mathbb{E}^\ast \\
	\mathcal{H}(p, f) &= \mathcal{H}(p) - I(w;p) + \mathbb{E}_{w \sim Q(w \vert \mathcal{S})} \text{KL}[ p \Vert f ] \notag \\
	&= \lambda \cdot \mathcal{H}(p) + (1-\lambda) \cdot \mathcal{H}(p) - I(w;p) + \mathbb{E}
\end{align}
We would like to simplify the term $\mathbb{E}^\ast - \mathbb{E}$. Note that we utilize the important property of KL divergence via log-sum inequality.
\begin{align*}
	\text{KL}(p^\ast \Vert f) &= \sum (\lambda p_1 + (1-\lambda) p_2)  \log \dfrac{\lambda  p_1 + (1-\lambda)  p_2}{\lambda  f + (1-\lambda) f} \\
	&\leq \lambda \cdot \text{KL} (p \Vert f) + (1-\lambda) \cdot \text{KL} (u \Vert f)
\end{align*}
Use the same trick on $\text{KL}(p \Vert f)$, and we could get
\begin{align*}
	\mathbb{E}^\ast - \mathbb{E} \leq (1-\lambda) \cdot [\text{KL}(u \Vert f) - \text{KL}(p \Vert f)]
\end{align*}
Therefore
\begin{align*}
	\mathcal{H}(p^\ast, f) - \mathcal{H}(p, f) &= (1-\lambda) \big( \mathcal{H}(p_2) - \mathcal{H}(p) \big) + \mathcal{Q} + \mathbb{E}^\ast - \mathbb{E} \\
	&\leq (1-\lambda) \cdot [\mathcal{H}^\ast + \mathcal{R}] + \mathcal{Q}
\end{align*} 
Here, $\mathcal{H}^\ast = \mathcal{H}(u) - \mathcal{H}(p)$ is always semi-positive and $\mathcal{R}=\text{KL}(u \Vert f) - \text{KL}(p \Vert f)$.
The difference of the two entropy could also be $(1-\lambda) \cdot \mathcal{H}(u,f) - \mathcal{H}(p,f)$ and then we could complete the proof, i.e., $\mathcal{Q} \geq 0$.
\qedd

\subsection{Proof of \cref{pro:1}}
\begin{pro}
	Let $\ell_{sce}, \ell_{rce}$ be the symmetric and reverse cross entropy loss function respectively and $\gamma$ represents their summation, i.e., $\ell_{sce}+\ell_{rce}=\gamma$. When $\gamma \to 1$, then our methods can also be written as:
	\begin{align*}
		\ell_{sglr} = \ell_{sce} - \alpha \cdot \ell_{rkl}
	\end{align*}
	where $\ell_{rkl}$ denotes the reverse KL divergence between labels and model predictions, \textit{i.e.}, $ D_{\text{KL}}(p \, || \, q)$.
	\label{appendix:pro:1}
\end{pro}
\textbf{\textit{Proof.}}\,
Formally, the conventional SCE loss can be written as:
\begin{align*}
	\ell_{sce} = \alpha \cdot \ell_{ce} + \beta \cdot \ell_{rce}
\end{align*}
Note that $\alpha,\beta=1$ is a special case of this form. We can still let $\gamma \to 1$, then
\begin{align}
	\begin{split}
			\ell_{sce} = \alpha \cdot \ell_{ce} + (1-\alpha) \cdot \ell_{rce}
	\end{split}
\end{align}
We take a closer look at the self-guided soft label and write $p(k;x)$ as $p(k)$ for simplicity \footnote{We omit the adversarial knowledge and historical average prediction temporarily without loss of generality.}. 
\begin{align}
	\begin{split}
		& q^\prime(k) = (1-\alpha) \cdot q(k) + \alpha \cdot p(k) \\
		\mathcal{H}(q^\prime (k),\,p(k)) = &-\sum_{k=1}^K (1-\alpha) \cdot q(k) \cdot \log p(k)  \\
		& + \alpha \cdot p(k) \cdot \log p(k) \\
		= &(1-\alpha) \cdot \mathcal{H}(q,\,p) + \alpha \cdot \mathcal{H}(p)
	\end{split}
\end{align}
where $q(\cdot)$ is the ground truth distribution over the labels and $\mathcal{H}(\cdot)$ denotes the cross entropy loss. Recall that the Kullback-Leibler Divergence could be dubbed as information gain, \textit{i.e.}, $D_{KL}(p \,||\, q) = \mathcal{H}(p,\,q) - \mathcal{H}(p)$. As $\gamma\to 1$, then the cross entropy loss of our method can also be written as:
\begin{align}
	\begin{split}
			\mathcal{H}(q^\prime(k), \,p(k)) = &(1-\alpha)\cdot \mathcal{H}(q,\,p) + \alpha \cdot \mathcal{H}(p,\,q) \\
			&- \alpha \cdot D_{KL}(p \, ||\, q) \\
			= &\ell_{sce} + \alpha \cdot \ell_{rkl}
	\end{split}
\end{align}
\qedd

\subsection{Proof of \cref{pro:2}}
\begin{pro}
	Some KD methods, which minimize the distance of the feature map between the teacher and student model, belong to the family of our method. Let $p_t$ be the prediction of the teacher model and then the KD could also be written as $\ell_{\text{KD}} = \mathbb{E}_{\widetilde{q}} \, \left[  -\log p  \right] = \mathcal{H}(\widetilde{q},\,p)$, where $\widetilde{q} = (1-\alpha) \cdot q + \alpha \cdot p_t$.
	\label{appendix:pro:2}
\end{pro}
\textbf{\textit{Proof.}}\,
Firstly, for the self-guided soft label $q^\prime = (1-\alpha) \cdot q + \alpha \cdot p$, if we replace the self-prediction $p$ with the knowldege of teacher model $p_t$, we have
\begin{align}
	\widetilde{q} = (1-\alpha) \cdot q + \alpha \cdot p_t
\end{align}
We utilize the special form $\widetilde{q}$ and have
\begin{align}
	\begin{split}
		\mathcal{H}(\widetilde{q} ,\, p) &= - \sum \widetilde{q} \cdot \log p \\
		&= -(1-\alpha) \sum q \cdot \log p - \alpha \sum p_t \cdot \log p \\
		&= (1-\alpha)\cdot \mathcal{H}(q, \, p) + \alpha \cdot \mathcal{H}(p_t, \,p)
	\end{split}
\end{align}
We here apply KL equality again
\begin{align}
	D_{\text{KL}}(p_t \,||\, p) = \mathcal{H}(p_t, \,p)-\mathcal{H}(p_t)
\end{align}
Note that $\mathcal{H}(p_t)$ represents the entropy of teacher prediction. When teacher is fixed, $\mathcal{H}(p_t)$ is a constant so that we can miss it during loss minimization. Then the loss of special soft label can be written as:
\begin{align}
	\begin{split}
		\mathcal{H}(\widetilde{q} ,\, p) &= (1-\alpha) \cdot \mathcal{H}(q, \, p) + \alpha \cdot D_{\text{KL}}(p_t \,||\, p)
	\end{split}
\end{align}
Some KD methods minimize this loss function $(1-\alpha) \cdot \mathcal{H}(q, \, p) + \alpha \cdot D_{\text{KL}}(p_t \,||\, p)$ and belong to the family of our method.
\qedd

\subsection{Proof of \cref{thm:1}}
\begin{theorem}
	In a $K$-class classification problem, $\widetilde{\ell}$ is noise-tolerant under symmetric or uniform label noise if noise rate $\eta < 1- \frac{1}{K}$. And if $R(f^\ast)=0$, $\widetilde{l}$ is also noise-tolerant under asymmetric or class-dependent label noise when noise rate $\eta_{y,k} < 1- \eta_y$ with $\sum_{i \neq y} \eta_{y,i}=\eta_y$, then
	\begin{align*}
		R_\mathcal{S}^\eta(f^\ast)-  R_\mathcal{S}^\eta(f) \simeq(1 - \frac{\eta K}{K-1})(R_\mathcal{S}(f^\ast) - R_\mathcal{S}(f)) \leq 0
	\end{align*}
\end{theorem}
\textbf{\textit{Proof.}}\,
\textit{For symmetric noise:}
\begin{align*}
	R_\mathcal{S}^{\eta}(f)  = &\mathbb{E}_{x, y} \widetilde{\ell}(x,y)=\mathbb{E}_{x} \mathbb{E}_{y \mid x} \mathbb{E}_{\eta \mid x, y} \widetilde{\ell}(x,y) \\
	 = & \mathbb{E}_{x} \mathbb{E}_{y \mid x}\left[(1-\eta) \widetilde{\ell}(x,y)+\frac{\eta}{K-1} \sum_{k \neq y}^{K} \widetilde{\ell}(x,k)\right] \\
	 = & (1-\eta) R_\mathcal{S}(f) \\
	   & +\frac{\eta}{K-1}\left(\frac{1}{N}\sum_{i=1}^N \sum_{k=1}^{K} \widetilde{\ell}(x_i,k)-R_\mathcal{S}(f)\right) \\
	 = & R_{\mathcal{S}}(f)\left(1-\frac{\eta K}{K-1}\right)- \eta \cdot G(f)
\end{align*}
where $G(f)=-\frac{1}{N(K-1)}\sum_{i=1}^N f(x_i) \cdot \log f(x_i)$ depends on the performance of classifiers. Then we are more preoccupied with the magitude of the difference between $G(f^\ast)$ and $G(f)$. Considering that the global minimum is adept at fitting the training data with nearly zero loss and we can make an appropriate assumption that $G(f^\ast)$ tends to be zero. Besides, we calculate the worst-case of the classifier, which randomly guesses the given data, and the entropy achieve maximum value. The difference could be approximated to $C=\eta \cdot (1-\frac{1}{K})\cdot 0.1$ and thus,
\begin{align*}
	R_\mathcal{S}^\eta(f^\ast) - R_\mathcal{S}^\eta(f) = (1 - \frac{\eta K}{K-1})(R_\mathcal{S}(f^\ast) - R_\mathcal{S}(f)) + C
\end{align*}
since $f^\ast$ is the global minimum of $R_\mathcal{S}(f)$ undert noise free data and the constant $C$ could be approximately igore as the small magnitude. This proves that $f^\ast$ is also the the global minimizer of $R_\mathcal{S}^\eta(f)$ and our method is nearly noise-tolerant.

\textit{For asymmetric or class-dependent noise}, $1-\eta_y$ is the probability of a label being correct (\textit{i.e.}, $k=y$), and the noise condition $\eta_{y,k}<1-\eta_y$ generally states that a sample $x$ still has the highest probability of being in the correct class $y$, though
it has probability of $\eta_{y,k}$ being in an arbitrary noisy (incorrect) class $k\neq y$. Following the symmetric case, here we set $C=\max \sum_{i=1}^{N}\sum_{k=1}^{K} \widetilde{\ell}(x_i, k)$ and thus,
\begin{align}
	R^{\eta}(f) = &\mathbb{E}_{x, y} \widetilde{\ell}(x, y)=\mathbb{E}_{x} \mathbb{E}_{y \mid x} \mathbb{E}_{y \mid x, y} \widetilde{\ell}(x, y)  \\ 
	 = & \mathbb{E}_{x} \mathbb{E}_{y \mid x}\left[\left(1-\eta_{y}\right) \widetilde{\ell}(x, y)+\sum_{k \neq y} \eta_{y k} \widetilde{\ell}(x, k)\right] \\
	 = & \mathbb{E}_{x, y}\left[\left(1-\eta_{y}\right)\left(\sum_{k=1}^{K} \widetilde{\ell}(x, k)-\sum_{k \neq y} \widetilde{\ell}(x, k)\right)\right] \notag{} \\
	   & + \mathbb{E}_{x, y}\left[\sum_{k \neq y} \eta_{y k} \widetilde{\ell}(x, k)\right] \\
	 = & \mathbb{E}_{x, y}\left[\left(1-\eta_{y}\right)\left(C-\sum_{k \neq y} \widetilde{\ell}(x, k)\right)\right] \notag{} \\
	   & + \mathbb{E}_{x, y}\left[\sum_{k \neq y} \eta_{y k} \widetilde{\ell}(x, k)\right] \\
	 = & C \cdot \mathbb{E}_{x, y}\left(1-\eta_{y}\right)\notag{} \\
	   & - \mathbb{E}_{x, y}\left[\sum_{k \neq y}\left(1-\eta_{y}-\eta_{y k}\right) \widetilde{\ell}(x, k)\right] .
	\label{eq:asymm}
\end{align}

Since $\eta_{y,k} < 1- \eta_y$ and $f_{\eta}^\ast$ is the minimizer of $R^\eta(f)$, $R^\eta(f^\ast_\eta) - R^\eta(f^\ast)\leq 0$. So, from \cref{eq:asymm},
\begin{align}
	\mathrm{E}_{x, y}\left[\sum_{k \neq y}\left(1-\eta_{y}-\eta_{y k}\right)(\widetilde{\ell}\left(f^{*}(\mathbf{x}), k\right)-\widetilde{\ell}\left(f_{\eta}^{*}(\mathbf{x}), k\right))\right] \leq 0 .
\end{align}
This proves that under asymmetric noise setting $f^\ast$ is also the the global minimizer of $R_\mathcal{S}^\eta(f)$ and our method is noise-tolerant.
\qedd

\section{More results}\label{appendix:results}


\begin{figure}[h]
	\centering
	\includegraphics[width=\linewidth]{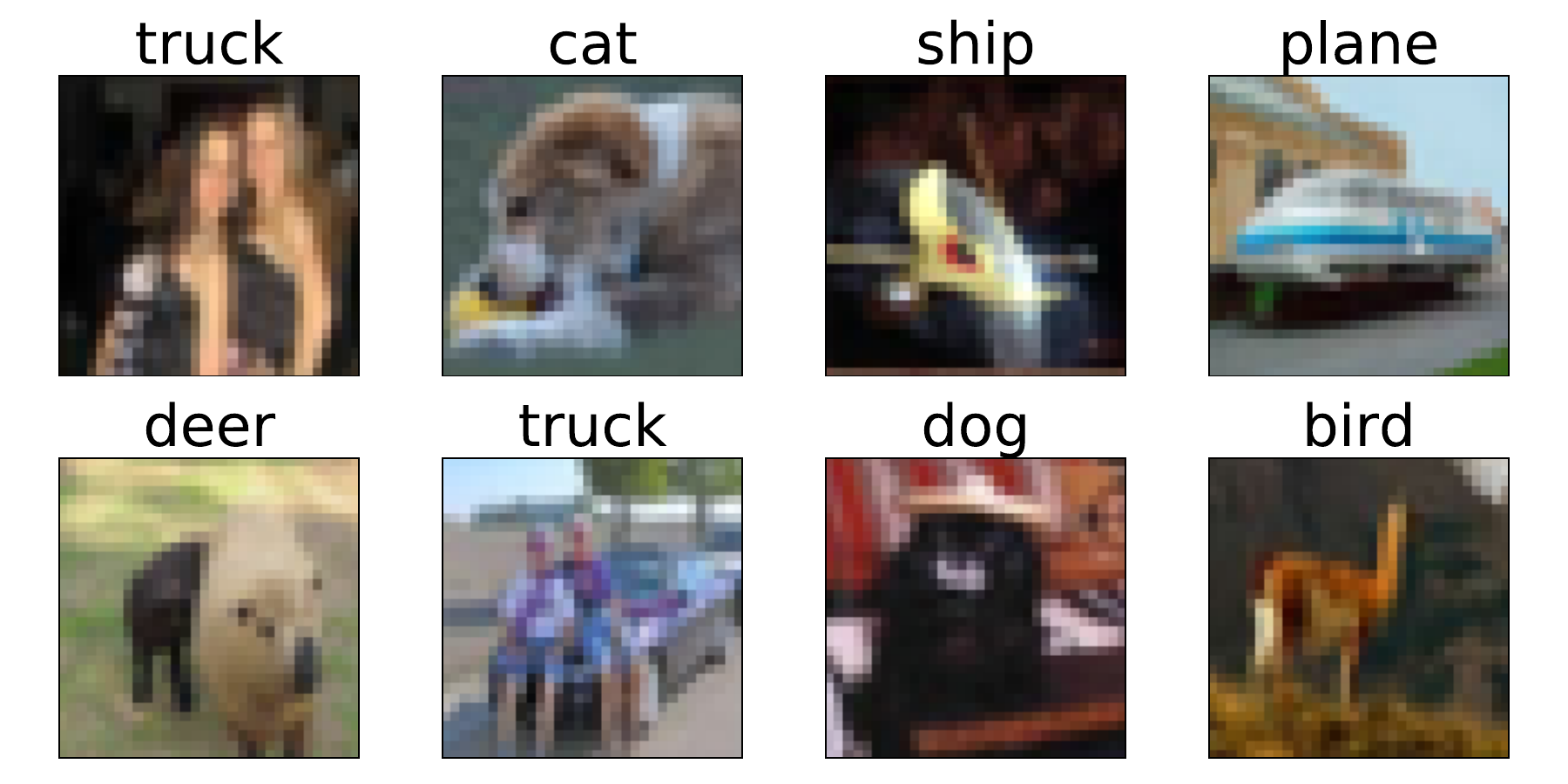}
	\caption{Label noise in CIFAR-10 training dataset.}
	\label{fig:ln}
\end{figure}
\subsection{Label noise in common datasets}
As reported in the study in \cite{benign}, label noise is surprisingly common in benchmark datasets such as CIFAR-10, which is presented in \cref{fig:ln}. We note that some examples are mislabelling (\textit{e.g.}, dog labeled with cat in \cref{fig:ln}). Such erroneously annotated samples make it hard for models to learn a good decision boundary. Unsurprising, feeding the model with such noisy labels would inevitably exacerbate the problem of robust overfitting and leads to {over confidence.} 

\begin{figure}[h]
	\centering
	\includegraphics[width=\linewidth]{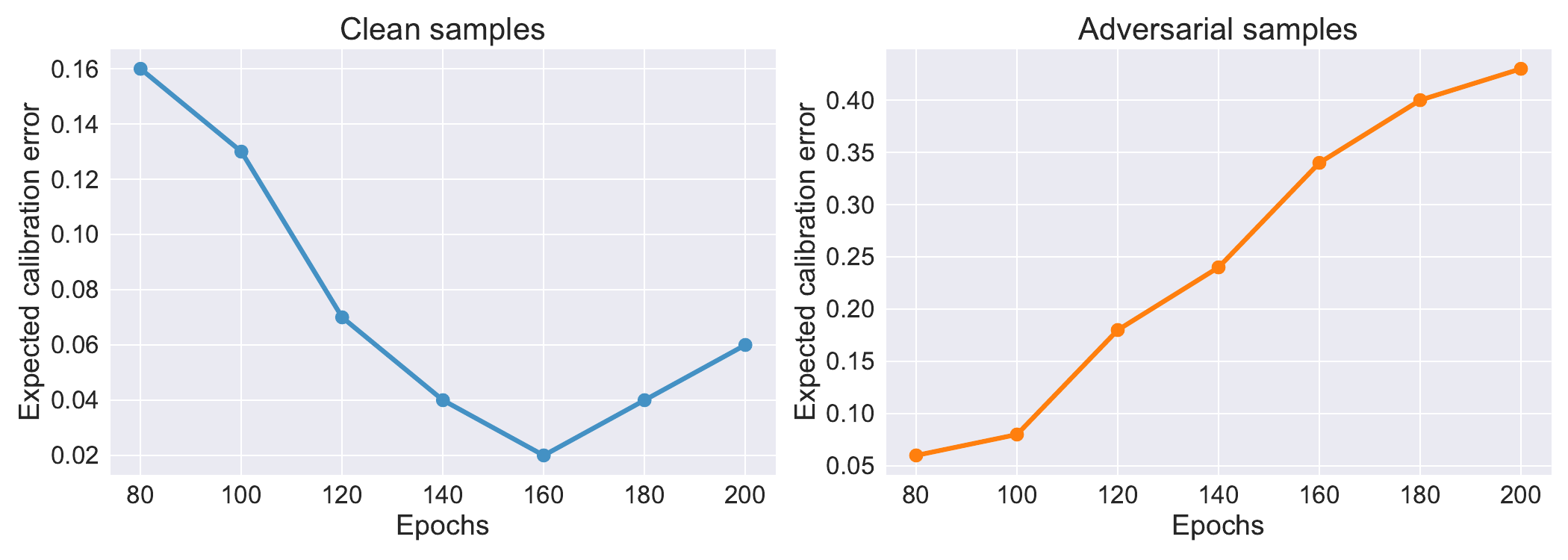}
	\caption{Expected calibration error for adversarial training on clean and adversarial samples. The learning rate is decayed by a factor of 0.1 at 100-th and 150-th.}
	\label{fig:ece}
\end{figure}
\begin{figure}[h]
	\centering
	\includegraphics[width=\linewidth]{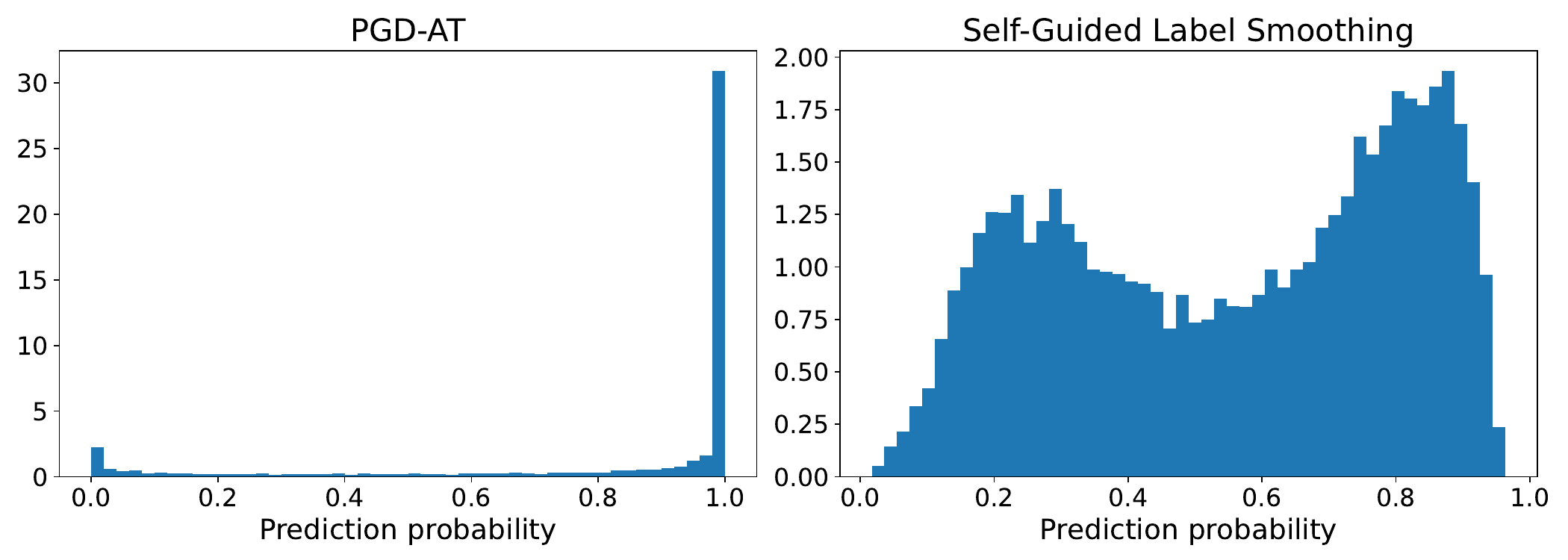}
	\caption{Samples density \textit{w.r.t.} the prediction probabilities (the softmax outputs on the labeled class).}
	\label{appendix:density}
\end{figure}

\subsection{More results about calibration}
A ideally trustworthy and reliable model ought to provide high confidence prediction on correct category and contrariwise. So, we adopt expected calibration error (ECE) for a model $f$ with $0<m<\infty$, as suggested in \cite{ece}.
\begin{align*}
	\text{ECE}_{p} = \mathbb{E} \left[ \mid \hat{z} - \mathbb{E}[ \delta_{y,\hat{y}} \hat{z}]\mid ^m \right]^{\frac{1}{m}}
\end{align*} 
where $\delta_{i,j}$ is the Kronecker delta, which appears true if the variables are equal. We visualize the ECE during whole adversarial training procedure in \cref{fig:ece}. 

From \cref{fig:ece}, we can observe that vanilla adversarial training weakens the over-confident prediction on clean samples, thus achieving a good calibration than standard training. However, as the training progresses, the expected calibration error on adversarial samples shows a rapid ramp-up. It also indicates that the latter predition of the adversarially trained model is not well-calibrated and thus being not able to provide trustworthy knowledge.

\begin{figure*}[t]
	\centering
	\includegraphics[width=0.8\textwidth]{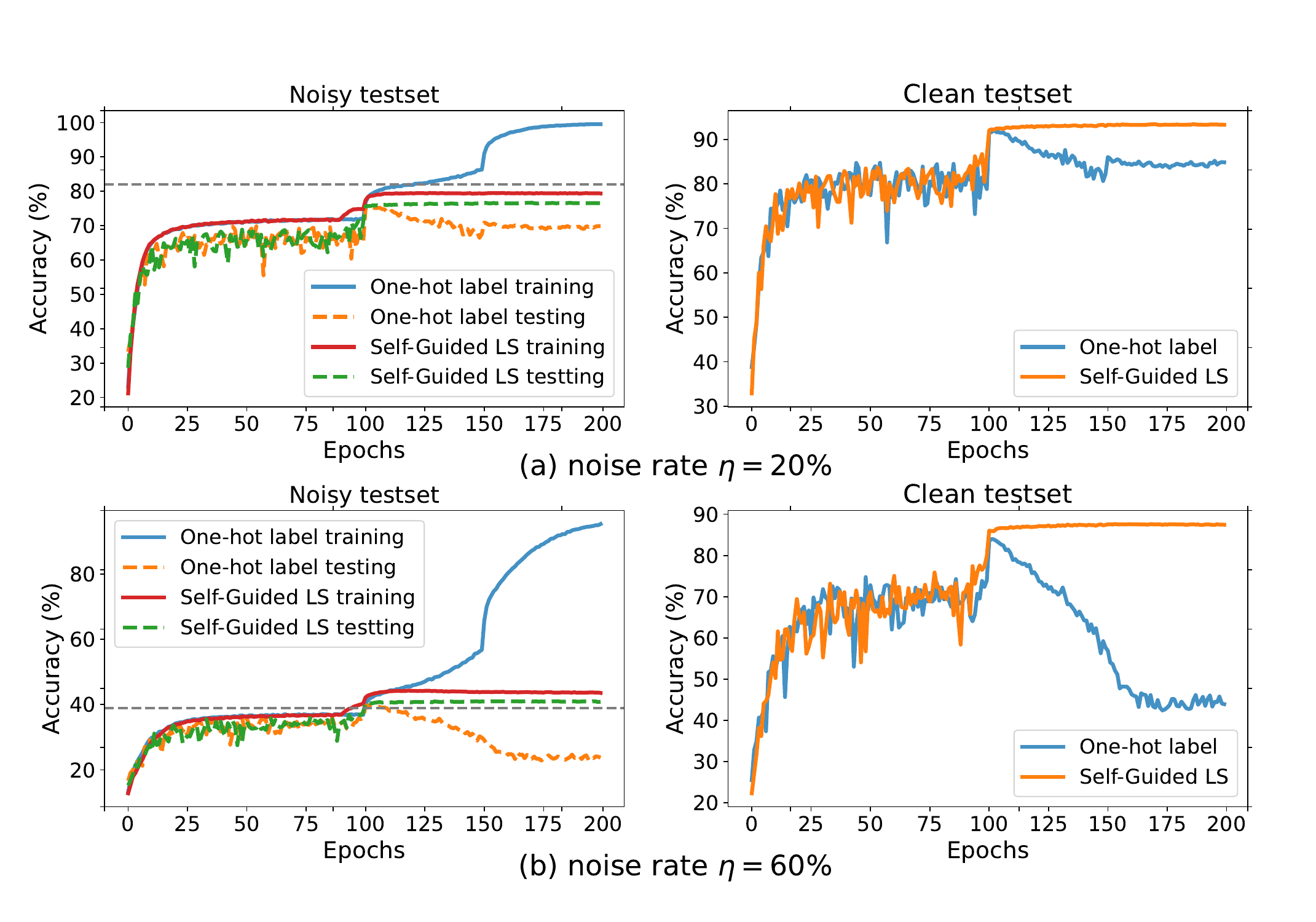}
	\caption{Test accuracy (\%) on clean and different noisy CIFAR-10 testset. The horizontal gray dashed line denotes the portion of correct labels.}
	\label{fig:noise_appendix}
\end{figure*}

\begin{table}[h]\footnotesize
	\centering
	\caption{Expected calibration error for PGD-AT and our proposed method on white-box (PGD-10) attack and black-box (square) attack.}
	\begin{tabular}{l c c c c c c}
		\toprule[1.5pt]
		& \multicolumn{3}{c}{White-box (PGD attack)} & \multicolumn{3}{c}{Black-box (Square attack)} \\
		& Best $\downarrow $  & Final  $\downarrow $ & Diff  $\downarrow $  & Best $\downarrow $  & Final  $\downarrow $ & Diff  $\downarrow $  \\
		\midrule[1pt]
		AT      & 0.18  & 0.43   & -0.25  & 0.07   &   0.39      &   -0.32      \\
		+SGLR & 0.11  & 0.10    & 0.01    & 0.20    & 0.22    & -0.02   \\
		\bottomrule[1.5pt]
	\end{tabular}
	\label{tab:ece}
\end{table}

Further, we also report the expected calibration error for PGD-AT and our method on both best and final checkpoint under different attacks. From \cref{tab:ece}, we can observe that our method effectively decrease the calibration error and thus alliviate the over confident prediction. Additionally, we plot the sample density \textit{w.r.t.} predictions on the labeled class in \cref{appendix:density}. We reach the same conclusion that our proposed method reduce the overconfidence mostly and thus achieve good generalization.

\begin{figure}[h]
	\centering
	\includegraphics[width=\linewidth]{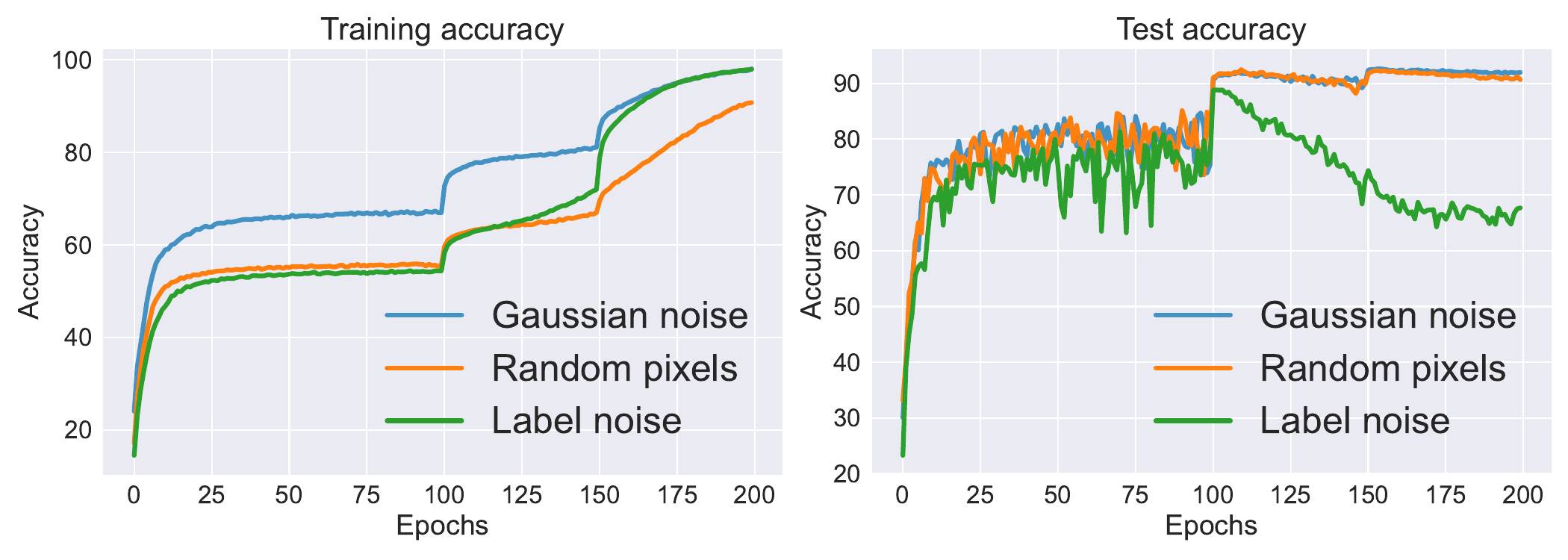}
	\caption{Training and test accuracy (\%) under different noise type in CIFAR-10 training dataset.}
	\label{fig:noise_type_appendix}
\end{figure}

\subsection{More experiments under different noise settings}
As reported in \cite{rice}, robust overfitting has been prevalent across various datasets and models. However, it may not occur while the strength of attack is relatively weak. We observe that smaller perturbation does not lead to a double-descent test accuracy and even obtains an overall increase, while large perturbation induces robust overfitting. 

As discussed in \cref{sec:2}, robust overfitting is similar to label noise in standard training to some extend. Namely, when the noise rete ramps up, it is on the verge of occuring with double descent curves. 

From \cref{fig:noise_type_appendix}, we can obverve that other noise types (like Gaussian and random pixels), even with extremely large perturbation, do not incur overfitting. The training accuracy and test accuracy consistently increase as the learning rate decayed though the clean test accuracy oscillates in its infancy. 

Besides, we provide the training procedure of hard label and self-guided soft label over different noise rate on CIFAR-10 dataset in \cref{fig:noise_appendix} and \cref{tab:appendix_noise} . It is worth noting that self-guided soft label constantly narrow the training and testing gap as the noise rate ramps up, while the hard label still memorizes the training data and eventually leads to bad testing accuracy.

\begin{table}[h]
	\caption{Evaluating different label strategies at various noise rates.}
	\begin{tabular}{c c c c c}
	\toprule[1.5pt]
	Rate           & 0\%          & 20\%           & 60\%         & 80\%         \\
	\midrule
	Hard Label  & 26.8 & 37.8    & 62.1  & 81.6  \\
	Soft Label         & 26.4 & 35.5   & 59.5  & 80.4 \\
	\midrule
	Self-Guided Soft Label        & 25.5 & 33.8    & 51.3  & 78.9  \\
	\bottomrule[1.5pt]
	\end{tabular}
	\label{tab:appendix_noise}
  \end{table}

\begin{table*}[t]\small
	\centering
	\caption{Performance (\%) of PGD-AT and our proposed method against different black-box attacks.}
	\begin{tabular}{l l l l l l l}
		\toprule[1.5pt]
		\multirow{2}{*}{ResNet-34 $\to$ ResNet-18}	& \multicolumn{3}{c}{PGD-10} & \multicolumn{3}{c}{$\text{CW}_{\infty}$} \\
		& Best   & Final  & Diff  & Best  & Final  & Diff  \\
		\midrule[1pt]
		AT      & 63.9   & 64.9   & -1.0    & 72.5  & 70.1     & 2.4  \\
		AT+SGLR & 64.0     & 64.1   & -0.1  & 72.7  & 72.9   & -0.2  \\
		\midrule[1pt] \Bstrut
		\multirow{2}{*}{ResNet-50 $\to$ ResNet-18} & \multicolumn{3}{c}{PGD-10} & \multicolumn{3}{c}{$\text{CW}_{\infty}$} \\
		& Best   & Final  & Diff  & Best  & Final  & Diff  \\
		\midrule[1pt]
		AT      & 81.3   & 82.7   & -1.4  & 83.1  & 81.6   & 1.5   \\
		AT+SGLR & 80.9   & 80.9   & 0.0     & 83.0    & 82.8   & 0.2   \\
		\bottomrule[1.5pt]
	\end{tabular}
	\label{appendix:black}
\end{table*}
\begin{table*}[t]\small
	\centering
	\caption{Clean accuracy and robust accuracy (\%) against white-box attacks of networks. All threat models are under $\ell_\infty$ norm with $\epsilon=8/255$. The bold indicates the improved performance achieved by the proposed method.}
	\begin{tabular}{l c c c l l l l l l}
		\toprule[1.5pt]
		\multicolumn{1}{c}{\multirow{2}{*}{Method}} & \multicolumn{3}{c}{Natrural Accuracy} & \multicolumn{3}{c}{PGD-20} & \multicolumn{3}{c}{AutoAttack} \\
		\multicolumn{1}{c}{}                        & Best       & Final       & Diff $\downarrow $      & Best    & Final   & Diff $\downarrow $  & Best     & Final     & Diff $\downarrow $    \\ \hline 
		\multicolumn{10}{c}{\cellcolor{emph} ResNet-18} \TBstrut \\ \hline \TBstrut
		AT                                          &  80.7       &     82.4        &     -1.6      &    50.7     &   41.4      &   9.3     &   47.7       &    40.2       &   7.5        \\
		{\textbf{+SGLR}}                                 &   {\textbf{82.9}}         &    {\textbf{83.0}}           &   {\textbf{0.1}}           &   {\textbf{56.4}}        &    {\textbf{55.9}}       &   {\textbf{0.5}}       &    {\textbf{51.2}}        &       {\textbf{50.2}}      &   {\textbf{1.0}}        \\ 
		\hline
		\multicolumn{10}{c}{\cellcolor{emph} ResNet-34-10} \TBstrut \\ \hline \TBstrut
		AT                                          &     \textbf{87.6}        &    86.4         &    1.2       &    55.9     &   50.2      &   5.7     &   51.2       &       45.6    &   5.6        \\
		{\textbf{+SGLR}}                                 &   {87.4}         &    {\textbf{87.2}}           &   {\textbf{0.2}}           &   {\textbf{59.5}}        &   {\textbf{58.0}}       &   {\textbf{1.5}}       &    {\textbf{54.3}}        &       {\textbf{52.3}}      &   {\textbf{2.0}}        \\
		\bottomrule[1.5pt]       
	\end{tabular}
	\label{appendix:tab:2}
\end{table*}

\subsection{More experiments about black-box attacks and large model architecture}
Additionally, we present evaluations on black-box attacks, \textit{i.e.}, adversarial examples generated from a different model (typiclally from a larger model), in \cref{appendix:black}. Here, we test ResNet-18 trained under PGD-AT and our proposed method with crafted adversarial examples from ResNet-34 and ResNet-50 trained with vanilla AT. Results in \cref{appendix:black} demonstrate that our method indeed close the gap between best and final checkpoint. These results not only show that our method does not suffer from the gradient obfuscation but also show that our method is effective in black-box attack settings.

\begin{figure}[h]
	\centering
	\includegraphics[width=\linewidth]{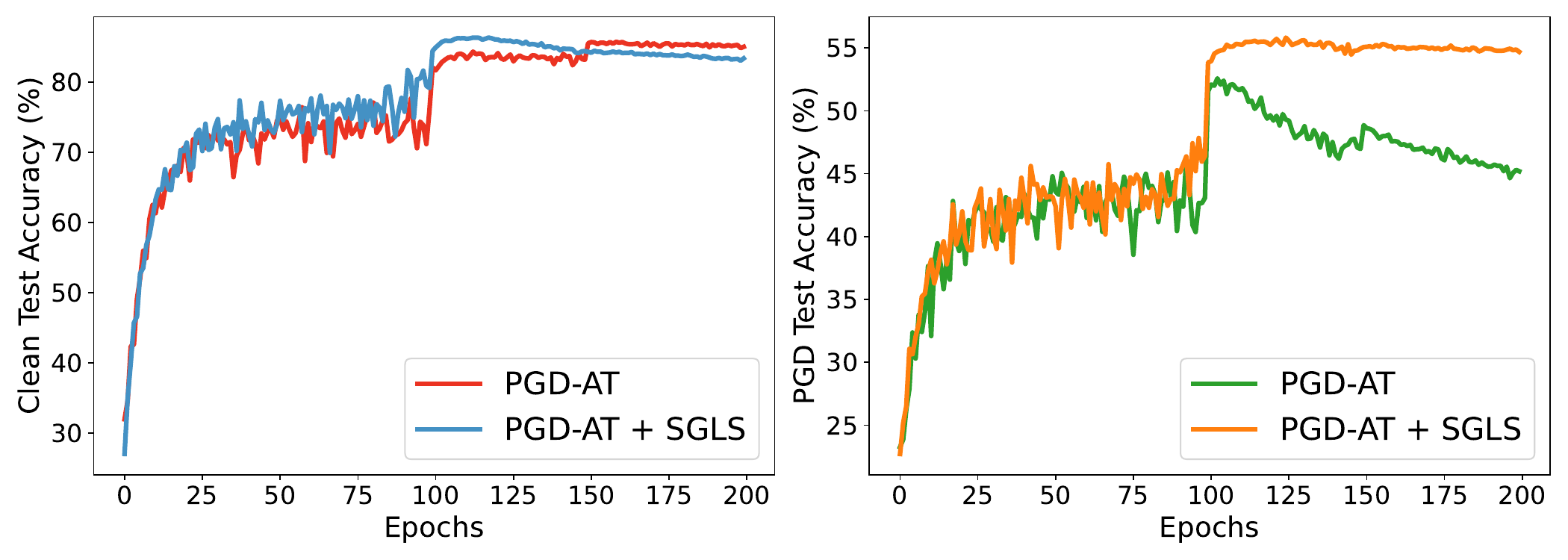}
	\caption{Result of training and testing accuracy over epochs for ResNet-18 trained on CIFAR-10.}
	\label{fig:ro_appendix}
\end{figure}

Furthermore, performance against various white-box attacks for large model architecture are shown in \cref{appendix:tab:2}. We similarly found that combing PGD-AT with our method could achieve superior performance even under strong autoattack. Notably, our method can largely reduce the gap between the best and final accuracies and thus effectively prevent robust overfitting.

\subsection{Different learning rate strategies}
\begin{figure}[h]
	\centering
	\includegraphics[width = \linewidth, trim=0cm 0cm 0cm 0.6cm, clip]{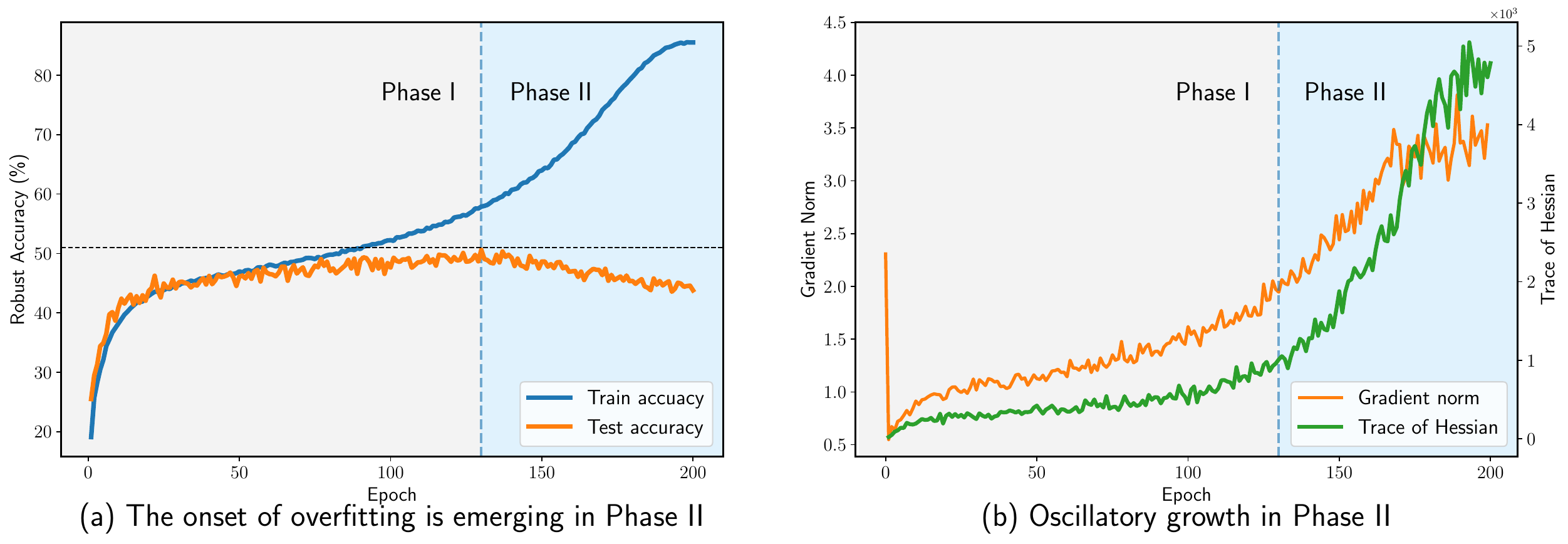}
	\caption{Different phases of training.}
	\label{fig:cosine}
  \end{figure}

The staircase learning rate schedule (piece-wise) is typiclally applied in adversarial training, which may have negative influence in obtaining robust models. In \cref{fig:cosine}, we plot the \textit{test robust accuracy}, \textit{gradient norm} and \textit{trace of hessia}n, which is widely used to measure the sharpness. As shown, training with cosine learning rate schedule yields smoother curves compared to that of the piece-wise learning rate schedule. 
Note that it does not prevent the widening generalization gap and robust overfitting, only influencing the \textbf{duration of the Stationary Stage}. The {\color{comparison}{green}} in \cref{fig:cosine} supplements
this with the trace of hessian, to better illustrate the characteristics between the two stages.

\section{Algorithm}
\begin{algorithm}
	\caption{Self-Guided Label Smoothing}
	\begin{algorithmic}
		\REQUIRE Total Epoch $N$, Neural Network $f_\theta$ with parameters $\theta$, Training Set $\mathcal{D} = \{ (x_j,y_j) \}$.
		\STATE $\widetilde{p_t} = 0$
		\FORALL{$\text{epoch}=1, \cdots, N$}
		\FORALL{$(x_j,y_j) \in \mathcal{D}$}
		\STATE {\color{blue}{* Inner Maximization to update $\delta$}}
		\STATE \quad $\delta \gets \arg\max \limits_{\Vert \delta \Vert_p \leq \epsilon} \ell(f(x),y)$
		\STATE {\color{blue}{* Outer Minimization to update $\theta$}}
		\STATE \quad $\widetilde{f}(x,x^\prime;\theta_t) = \lambda \cdot f(x;\theta) + (1-\lambda)\cdot 
		f(x^\prime;\theta)$ \, \COMMENT{\# Self-Guided Label Refinement}
		\STATE \quad $\mathbf{y} = r \cdot \widetilde{p_t} + (1-r) \cdot \mathbf{y}_{hard}$
		\STATE \quad $\widetilde{p_t} = \alpha \cdot \widetilde{p}_{t-1} + (1-\alpha) \cdot \widetilde{f}(x,x^\prime;\theta_t)$ \COMMENT{\# Consensus of the self-distilled models}
		\STATE \quad $\ell_{sglr} = \ell(f(x), \widetilde{p_t})$ 
		\STATE \quad $\theta \gets \theta - \eta \cdot (\nabla_\theta \widetilde{L})$
		\ENDFOR
		\ENDFOR
	\end{algorithmic}
	\label{al}
\end{algorithm}